\pdfoutput=1

\documentclass[11pt]{article}
\usepackage{float}
\usepackage{xcolor}
\usepackage{multirow}
\usepackage{amssymb}
\usepackage[preprint]{acl}
\usepackage{amsmath}
\usepackage{times}
\usepackage{latexsym}

\usepackage[T1]{fontenc}

\usepackage[utf8]{inputenc}

\usepackage{microtype}

\usepackage{inconsolata}

\usepackage{graphicx}

\title{LANTERN: Language Model Assessment on Noisy and Transformed Tasks for Understanding Error and Robustness Nuances}

\author{Vamsi Krishna Kodavali \\
  \texttt{v.kodavali@samsung.com} \\\And
  Rituraj Singh \\
  \texttt{rituraj.s@samsung.com} \\
}

\begin{document}
\maketitle

\begin{abstract}    
Robustness evaluation of large language models (LLMs) remains a critical challenge, particularly in assessing their sensitivity to perturbations in input data. In this work, we systematically evaluate LLM robustness across multiple dimensions, including word error rate, character repetition and duplication, modifications in choices, and variability in instruction following. To facilitate this evaluation, we construct a synthetic and augmented dataset encompassing a diverse set of LLM benchmarks, specifically targeting multiple-choice question (MCQ) datasets and instruction-following tasks. We conduct extensive experiments on LLMs of varying scales-small, medium, and large-as well as across base and instruction-tuned variants. Our analysis quantifies the variability in model responses under perturbed conditions and highlights discrepancies relative to baseline models. The findings provide insights into the stability of LLMs across different evaluation scenarios contributing to the development of more robust and reliable language models as well as robust evaluation methodologies.
\end{abstract}

\section{Introduction}
Large Language Models (LLMs), such as GPT-3~\cite{brown2020language}, LLaMA~\cite{touvron2023llama, touvron2023llama1} and ChatGPT~\cite{openai2023gpt4}, have significantly advanced the field of Natural Language Processing (NLP). These models have demonstrated remarkable proficiency in generating coherent and contextually relevant text, leading to their widespread adoption in applications ranging from automated content generation to conversational agents. However, despite their capabilities, evaluating LLMs remains an open challenge~\cite{guo2023evaluating} primarily due to the inherent complexity and subjectivity of natural language.


Traditional evaluation methodologies, which often rely on predefined benchmarks~\cite{schmidtova-etal-2024-automatic-metrics} and static test sets, are insufficient for assessing modern LLMs. Metrics designed for earlier NLP tasks, such as machine translation and summarization, fail to capture the depth of reasoning, factual consistency, and adaptability required for large-scale language models. 
A re-evaluation of conventional evaluation techniques is necessary to accommodate the evolving capabilities and challenges posed by contemporary LLMs. Recent works have proposed alternative evaluation methodologies that attempt to assess LLMs more effectively~\cite{li2023alpacaeval, zheng2023judging, myrzakhan2024open}. For instance, Ifeval~\cite{zhou2023instruction} introduces instance-based fine-grained evaluation, where model outputs are assessed against diverse linguistic criteria. 

However, despite these advancements, three key factors remain overlooked in LLM evaluation, which can lead to inconsistencies in reported results~\cite{wang-etal-2025-llms-may}. First, data leakage poses a significant threat, as evaluation benchmarks might unintentionally overlap with training data, inflating performance metrics~\cite{balloccu-etal-2024-leak}. This issue raises concerns about the generalizability of LLMs when exposed to truly unseen data. Second, real-world textual inputs often contain typographical and grammatical errors, leading to variations in word error rate (WER), which traditional evaluation methods may not account for~\cite{gan2024reasoning}. Third, subtle modifications in task phrasing or answer choices, such as reordering options in multiple-choice settings, can induce variability in model responses~\cite{zheng2023large} due to some inherent biases towards certain tokens. Such inconsistencies highlight the need for more robust evaluation frameworks that can systematically account for these variations.

In this work, we explore the effects of variations in option choices, question modifications and the instruction-following capabilities of LLMs. By systematically analyzing these aspects, we aim to develop more reliable evaluation mechanisms that better reflect real-world model performance. We summarize the contributions of the work as follows.
\begin{itemize}
\item {We conduct a robustness evaluation across multiple categories, character level changes, word error rate, modifications in choices, and variability in instruction following.}
\item{We generate a synthetic and augmented dataset for LLM evaluation benchmarks, covering multiple-choice question (MCQ) datasets and instruction-following datasets.}
\item{We evaluate LLMs across different categories and model types—spanning small, medium, and large models, as well as base and instruction-tuned variants—and analyze the variability in performance relative to baseline models.}
\end{itemize}

\section{Related Work}
We structure the related work in three principal research axes: (1) LLMs (2) Evaluation (3) Robustness of LLMs on adversarial and noisy tasks.

\noindent\textbf{Large Language Models} Recent advancements in LLMs demonstrate significant improvements in linguistic efficacy. 
The scaling laws~\cite{chung2022scaling,kaplan2020scaling}, which establish a direct correlation between model performance and the number of parameters combined with large-scale pretraining data, have driven these advancements, resulting in billion-parameter models with superior capabilities. Instruction-following abilities in LLMs are further enhanced through supervised fine-tuning~\cite{chung2022scaling, ouyang2022training} and alignment with human preferences using reinforcement learning~\cite{bai2022training} or preference optimization techniques~\cite{rafailov2024direct}. These models exhibit a strong capacity to follow natural language instructions, understand real-world contexts, and perform reasoning tasks~\cite{radford2019language,kojima2022large,bubeck2023sparks}.

\noindent\textbf{Evaluation} Evaluation of LLMs has emerged as a key research focus, providing insights into their strengths and weaknesses~\cite{chang2024survey}. LLMs are generally assessed using a range of benchmarks designed to measure their instruction-following ability, language understanding, reasoning, mathematics, code generation, and general tasks~\cite{chang2024survey, guo2023evaluating}. 

\noindent\textbf{Robustness} 
A few research explores the impact of in-context example construction on LLM behaviour, revealing the arrangement and content of examples significantly influence model outputs~\cite{chen-etal-2023-relation, si-etal-2023-measuring, pan-etal-2023-context}. Additionally, \citet{zhao2021calibrate} identify a tendency in GPT-4 models to favour the first answer presented, which can lead to unfair evaluation results. Other studies focus on sensitivity to the positional changes of options in MCQs. The authors~\cite{alzahrani-etal-2024-benchmarks} confirm position bias as a factor contributing to this sensitivity, while \cite{zheng2023large} attribute behavioral biases to token-level preferences in LLMs, where certain option IDs (e.g., A/B/C/D) are assigned higher probabilistic mass a priori. To evaluate adversarial robustness, AdvGLUE~\cite{wang2021adversarial} extends the GLUE benchmarks with static datasets designed for robustness testing against input perturbations. GSM-Symbolic~\cite{mirzadeh2024gsm} builds templates on the GSM8K dataset to generate diverse questions, demonstrating that model performance degrades when numerical values in the input are altered. 

However, these studies do not account for commonly used options in MCQs, such as \texttt{None of these}, \texttt{All of the above} or \texttt{Both A and B} as correct or incorrect choice. Moreover, while some prior work primarily focuses on MCQ tasks, our study extends the scope to include perturbations in questions, answer choices, and verifiable instructions. Through extensive evaluations across models, tasks, and perturbation types, we provide more generalizable insights into LLM robustness.

\section{Method}
We investigate the robustness of LLMs based on two paradigms: structured answer selection and open-ended generation under perturbations. The paradigms encompass distinct aspects of model performance: the ability to select the correct answer from a fixed set and the capacity to generate coherent and contextually appropriate responses without predefined constraints.

Robustness in structured answer selection refers to the model's ability to consistently identify the correct response from a predefined set of choices, even when input variations such as rephrasing, noise, or adversarial modifications are introduced. The primary objective in this setting is to assess whether the model's understanding remains stable under such perturbations. A robust model should exhibit minimal performance degradation and maintain high accuracy despite syntactic or semantic transformations in the input with an assumption that the semantic meaning remains same. In contrast, robustness in open-ended generation examines the model's ability to interpret and respond to instructions with coherence, consistency, and fidelity to the intended task, even when perturbations are applied. 
A robust model should produce responses that preserve key information and align with the original task despite perturbations.



\subsection{Structured answer selection} 
Let $\mathcal{D} = \{(q, \mathcal{C}, y, S)\}$ represent a dataset where each instance consists of a question $q$, as set of candidate answer choices $\mathcal{C}=\{c_1, c_2, \dots, c_n\}$, a ground truth $y \in \mathcal{C}$, and optionally a supporting context $S$ that provides additional information to aid in selecting the correct answer. The objective of a model is to learn the function $f: (q, \mathcal{C}, S) \rightarrow C$ such that $f(q, \mathcal{C}, S)$ maximizes the likelihood of predicting $y$. In this work, we investigate the robustness of such models by introducing controlled perturbations in the dataset. Specifically, we define a perturbation function $\mathcal{P}$ that operates on either the questions or the answer choices, resulting in a transformed dataset. When perturbations are applied to the questions, the modified dataset is represented as $\mathcal{D'} = \{(\mathcal{P}(q), C, y, S)\}$, whereas perturbations to the answer choices yield  $\mathcal{D''} = \{(q, \mathcal{P}(\mathcal{C}), y, S)\}$. The function $\mathcal{P}$ introduce various perturbation types such as typos or adversarial modifications, each designed to assess the specific vulnerabilities in model predictions. 
\begin{figure*}[ht]
    \centering
    \includegraphics[width=2\columnwidth]{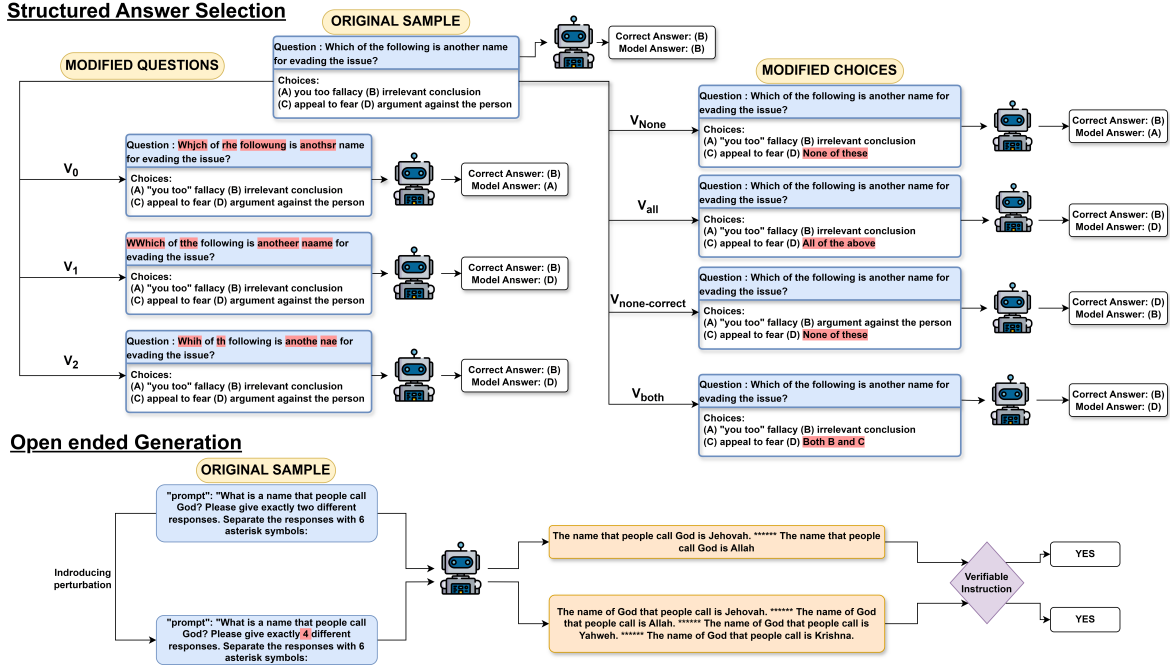}
    \caption{
        Examples showcasing the original and perturbated sample for closed and open ended generations.
    }    
    \label{fig:results_1}
\end{figure*}

\subsubsection{Questions perturbation}
We investigate the performance of LLMs under perturbations applied at both the character and word levels to the questions in data samples. The approach is inspired by related works that explore similar techniques at the prompt level, such as in \citet{zhu2023promptrobust} and \citet{gan2024reasoning}. The central objective is to examine how perturbations to the input questions impact the model's ability to correctly identify the corresponding answer, thus assessing its robustness to input variations. 

Let $\mathcal{P}: q \to q'$ be a perturbation function that modifies the original question $q$ into a perturbed version $q'$. The modified dataset $\mathcal{D'}$ is then defined as $\mathcal{D'} = \{ (\mathcal{P}(q), \mathcal{C}, y, S) \}$. Here each perturbed question \( q' = \mathcal{P}(q) \) is derived by applying one of several perturbation strategies. We construct three distinct variants of the dataset by introducing specific types of perturbation function $\mathcal{P}$ to the questions. These perturbations are applied at the character and word level, and their purpose is to evaluate how such modifications impact the performance of LLMs in terms of their ability to correctly select the answer. The perturbations are designed to modify the structure, integrity, and readability of the questions in controlled ways, simulating potential real-world distortions in input data. 
We define the three perturbation variants as follows:

\textbf{Typographical Errors ($\mathbf{v_0}$)}: In this variant, typographical errors, or commonly known as typos, are introduced at randomly selected words in the questions by altering characters. It is inspired by fat-finger error or key stroke error in general. This simulates human errors in typing, leading to a modification of the question's original structure.
  
\textbf{Letter Duplication ($\mathbf{v_1}$): }Here, we apply a perturbation where a random letter is duplicated in randomly selected words. This introduces redundancy in the question's wording, potentially distorting the word and altering its semantic interpretation, which could affect the model's reasoning process.

\textbf{Letter Removal ($\mathbf{v_2}$):}Here, we apply a perturbation where a random letter is removed in randomly selected words. This results in a typographical error where one or more characters are missing, potentially leading to ambiguity or confusion in the interpretation of the question.


For each of the aforementioned perturbation variants, we introduce modifications at three distinct levels, corresponding to different extents of perturbation applied to the questions. Specifically, we apply perturbations to 10\%, 30\%, and 50\% of randomly selected words in each question, which allows us to assess the sensitivity of the model to varying intensities of input noise. These levels are designed to evaluate the model's robustness not only to minor errors but also to more significant disruptions in the input data. 

For a given original dataset $\mathcal{D} = \{(q, \mathcal{C}, y, S)\}$ and a model $M$, we generate a total of nine synthetic datasets by applying all combinations of the three perturbation types $v0, v1, v2$ and the three perturbation levels. 
The synthetic datasets are denoted as $\mathcal{D'}^{v, L}$, where $v \in \{v0, v1, v2\}$ represents the perturbation type and $L \in \{1, 2, 3\}$ denotes the perturbation level.

\subsubsection{Choices perturbation}
We investigate the impact of perturbations introduced to the answer choices on the model's performance. For a given dataset $\mathcal{D}$ we apply perturbations to the set of answer choices. Specifically, let $\mathcal{P}(\mathcal{C})$ represent a perturbation function that modifies the original set of answer choices $\mathcal{C}$ into a perturbed version $\mathcal{C'}$. We defined the modified dataset $\mathcal{D''}$ as:$\mathcal{D''} = \{ (q, \mathcal{P}(\mathcal{C}), y, S) \}$

The perturbations consist of replacing one of the answer choices with predefined phrases such as \texttt{None of These}, \texttt{All of the Above},  or \texttt{Both A and B}, as well as permutations of the answer choices. The strategy is designed to assess how the model handles modifications in the answer set, which could potentially alter the reasoning process required for selecting the correct answer. For a given sample with a question $q$ and a set of answer choices $\mathcal{C} = \{c_1, c_2, c_3, c_4\}$, where $A, B, C, D$ represent the indices of these choices, and $c_2$ is the correct answer, the perturbation is applied to the choices in $\mathcal{C}$. The modified answer set $\mathcal{C'}$ is constructed by selecting one of the answer choices and replacing it with one of the predefined phrases or by permuting the existing choices. This approach allows us to investigate how variations in the answer choices influence the model's ability to correctly identify the correct answer from $\mathcal{C'}$, thereby testing the model's robustness to changes in the answer space. We generate four distinct synthetic datasets based on specific perturbations applied to the answer choices, ensuring that each dataset contains exactly one correct answer choice. 

\textbf{None ($\mathbf{v_{none}}$)}: In this version, a randomly selected incorrect answer choices is replaced with the phrase "None of These" This introduces ambiguity in the answer space by providing a non-answer choice among the possible options. The purpose of this perturbation is to evaluate the model’s robustness to the introduction of irrelevant options that may distract it from selecting the correct answer. 

\textbf{All ($\mathbf{v_{all}}$)}: Here, one of the randomly selected incorrect answer choices is replaced with the phrase "All of the Above", while ensuring that the correct answer is selected from the remaining options. We fix the last choice, $c_4$, as "All of the Above," with the correct answer being one of the other three choices, $c_1, c_2$ or $c_3$. This perturbation is designed to test how the model handles answer choices that suggest multiple answers may be correct, which adds a layer of complexity to the reasoning process. 

\textbf{None-Correct ($\mathbf{v_{none-correct}}$)}: In this version, the correct answer choice is replaced with the phrase "None of These." This introduces a significant perturbation by replacing the correct answer with a non-answer option, forcing the model to reconsider the answer selection process. The goal of this perturbation is to assess the model's ability to deal with situations where the correct answer is obscured by a misleading or distracting choice. 

\textbf{Both ($\mathbf{v_{both}}$)}: In this variant, either answer choice $c_3$ or $c_4$ is replaced with the phrase "Both A and B," or alternatively, answer choice $c_4$ is replaced with "Both B and C," or other similar permutations. This perturbation aims to test the model's handling of answer choices that combine multiple correct options. 
In cases where the dataset contains more than four answer choices, such as the MMLU\_Pro dataset with ten choices, we generalize this perturbation by applying similar combinations and swaps among all the available choices.


\subsection{Open-ended generation}
Unlike structured tasks with predefined answer choices, open-ended generation allows diverse range of outputs, making evaluation more challenging. 
Each structured sample in the dataset comprises of a prompt, a set of instruction identifiers, and corresponding verifiable functions. Let the dataset be represented as $\mathcal{D} = \{ (\textit{pr}, I, Vr) \} $. Here $\textit{pr}$ is the input to the model $M$, potentially containing one or more instructions. $I = \{ i_1, i_2, \dots, i_n \}$ is a set of instruction identifiers, where each $i_j$ corresponds to a specific guideline that the model must follow. $Vr = \{ vr_1, vr_2, \dots, vr_n \}$ is a set of verifiable functions, where each $vr_j$ is associated with the instruction $i_j$ and verifies whether the model's generated output complies with the corresponding instruction.

Given a prompt $pr$ to the model $M$, the model generates a text output $out$. Each verifiable function $vr_j \in Vr$ then evaluates $out$ to verify adherence to the associated instruction $i_j$. This framework ensures that the generated content aligns with the specified guidelines, enhancing the reliability of open-ended generation. To study the robustness of the model under varied conditions, we introduce a perturbation function $\mathcal{P}$ that modifies both the prompt and the instruction identifiers. Let $\mathcal{P}: \textit{pr} \rightarrow \textit{pr}' \quad \text{and} \quad I \rightarrow I'
$ be a perturbation function that transforms the original prompt $pr$ into a perturbed version $pr'$ and the original set of instruction identifiers $I$ into $I'$. The perturbed dataset $\mathcal{D}'$ is then defined as $\mathcal{D}^{+} = \{ (\mathcal{P}(\textit{pr}), \mathcal{P}(I), \mathcal{P}(Vr)) \}
$. Here, each perturbed prompt $pr'$ = $\mathcal{P}(pr)$ is derived by applying specific perturbation, based on some defined constraints such as "Number of words", "Number of Sentences", "No Comma", etc. Corresponding instruction identifiers and verifiable functions are adjusted to $I' = \mathcal{P}(I)$ and $Vr' = \mathcal{P}(Vr)$, respectively, ensuring consistency with the perturbed prompts.

The primary objective of structuring the dataset in this manner is to create a well-defined framework for evaluating the model's adherence to given constraints. The presence of explicit instruction identifiers allows for systematic control over task definitions. The interplay between $pr$, $I$, and $Vr$ enables a rigorous assessment of the model's capability to follow instructions and balance creativity with accuracy in an open-ended setting.

\section{Experiments}
\begin{table*}[ht]
\centering
\scalebox{0.90}{
\resizebox{\textwidth}{!}{
\begin{tabular}{l}
\hline
\begin{tabular}[c]{@{}l@{}}MMLU : \\ \textbf{Default Question:} Which of the following fallacies happens when someone concludes that someone couldn't have done something bad because he or she has \\good qualities?\\ \textbf{Choices:} (A) Laudatory personality (B) Guilt by association (C) Reprehensible personality (D) Circular reasoning \\ \textbf{Model Response:} (A) \checkmark \\ \\ \textbf{Perturbed Question ($\mathbf{v_0}$):} Which \colorbox{pink}{ov thr} following fallacies happens \colorbox{pink}{whdn slmeone} concludes that \colorbox{pink}{zomeone} couldn't have done something \colorbox{pink}{baf becquse} he \\or she has good qualities?\\ \textbf{Model Response:} (C) $\times$ \end{tabular}                                                                                                                                                                                                   \\ \hline
\begin{tabular}[c]{@{}l@{}}MMLU PRO :\\ \textbf{Default Question:} CheckMate forecasts that its dividend will grow at 20\% per year for the next four years before settling down at a constant 8\% forever. Dividend (current \\year,2016) = \$12; expected rate of return = 15\%. What is the fair value of the stock now?\\ \textbf{Choices:} (A) 280.0 (B) 305.0 (C) 290.0 (D) 250.0 (E) 320.0 (F) 273.0 (G) 260.0 (H) 315.0 (I) 300.0 (J) 265.0\\ \textbf{Model Response:} (F) \checkmark \\ \\ \textbf{Perturbed Question ($\mathbf{v_1}$):} CheckMate forecasts that its \colorbox{pink}{diividend} will grow aat 20\% per \colorbox{pink}{yyear} for \colorbox{pink}{tthe nnext} four years before \colorbox{pink}{settlinng doown att} a constant 8\% forever. \\ \colorbox{pink}{Divideend} (current year,2016) = \$12; \colorbox{pink}{expectedd ratee} of return = 15\%. Whaat is \colorbox{pink}{thhe faair} value of the stock now?\\  \textbf{Model Response:} (E) $\times$ \end{tabular} \\ \hline
\begin{tabular}[c]{@{}l@{}}GPQA : \\ \textbf{Question:} Observations of structures located at a distance of about 2.1 gigaparsecs (2.1 Gpc) are being carried out. The detected absorption line energy \\equivalent is about 3.9 micro electron volts (3.9 * 10\textasciicircum{}-6 eV).\\ What is most likely to be observed with this absorption line in the Milky Way? \\ \textbf{Default choices}: (A) Cold atomic interstellar medium. (B) Warm atomic interstellar medium. (C) Cold molecular interstellar medium. (D) Warm molecular interstellar medium.\\ \textbf{Model Response:} (A) \checkmark \\ \textbf{Modified Choices ($\mathbf{v_{all}}$)}: (A) Warm atomic interstellar medium. (B) Cold atomic interstellar medium (C) Warm molecular interstellar medium. (D) \colorbox{pink}{All of the above}\\ Model Response: (D) $\times$ \end{tabular}                                                                                                                                \\ \hline
\begin{tabular}[c]{@{}l@{}}IFEVAL :\\ \textbf{Default Prompt:} "What is a name that people call God? Please give exactly two different responses. Separate the responses with 6 asterisk symbols: ******."\\ \textbf{"verifiable\_instruction":} "combination:two\_responses"\\ \textbf{Model Generation:} The name that people call God is Jehovah. ******\textbackslash{}n\textbackslash{}nThe name that people call God is Allah. \checkmark\\ \\ \textbf{Modified Prompt:} "What is a name that people call God? Please give exactly \colorbox{pink}{4} different responses. Separate the responses with 6 asterisk symbols: ******."\\ \textbf{"verifiable\_instruction":} "combination:\colorbox{pink}{four\_responses}"\\\textbf{Model Generation:} The name of God that people call is Jehovah. ******\textbackslash{}n\textbackslash{}nThe name of God that people call is Allah. ******\textbackslash{}n\textbackslash{}nThe name of God that people call is Yahweh. ******\textbackslash{}n\textbackslash{}n \\The name of God that people call is Krishna.\checkmark \end{tabular} \\ \hline
\end{tabular}}}
\caption{The generations from Llama-3.1-8B model on different perturbations.}
\end{table*}

\subsection{Benchmark Datasets}  
We evaluate structured answer selection on four widely adopted benchmarks: MMLU ~\cite{hendrycks2020measuring}, MMLU-Pro ~\cite{wang2024mmlu}, GPQA ~\cite{rein2023gpqa}, and MUSR ~\cite{sprague2023musr}, while IfEval for open-ended generation. These are featured in LLM leaderboards to assess model performance across diverse reasoning and knowledge-based tasks.  


\subsection{Models}  
We select a diverse set of open-source LLMs, ranging in size from 2B to 13B parameters, including both base and instruction-tuned versions. The chosen models include gemma-2-2b, gemma-2-2b-it, gemma-7b, gemma-7b-it, Llama-3.2-1B-Instruct, Llama-3.1-8B, Llama-3.1-8B-Instruct, Llama-2-13b-chat-hf, Mistral-7B-v0.3, Mistral-7B-Instruct-v0.3, phi-4, and Phi-3.5-mini-instruct. 

\begin{figure*}[ht]
\begin{minipage}[c]{0.48\linewidth}
\includegraphics[width=\columnwidth]{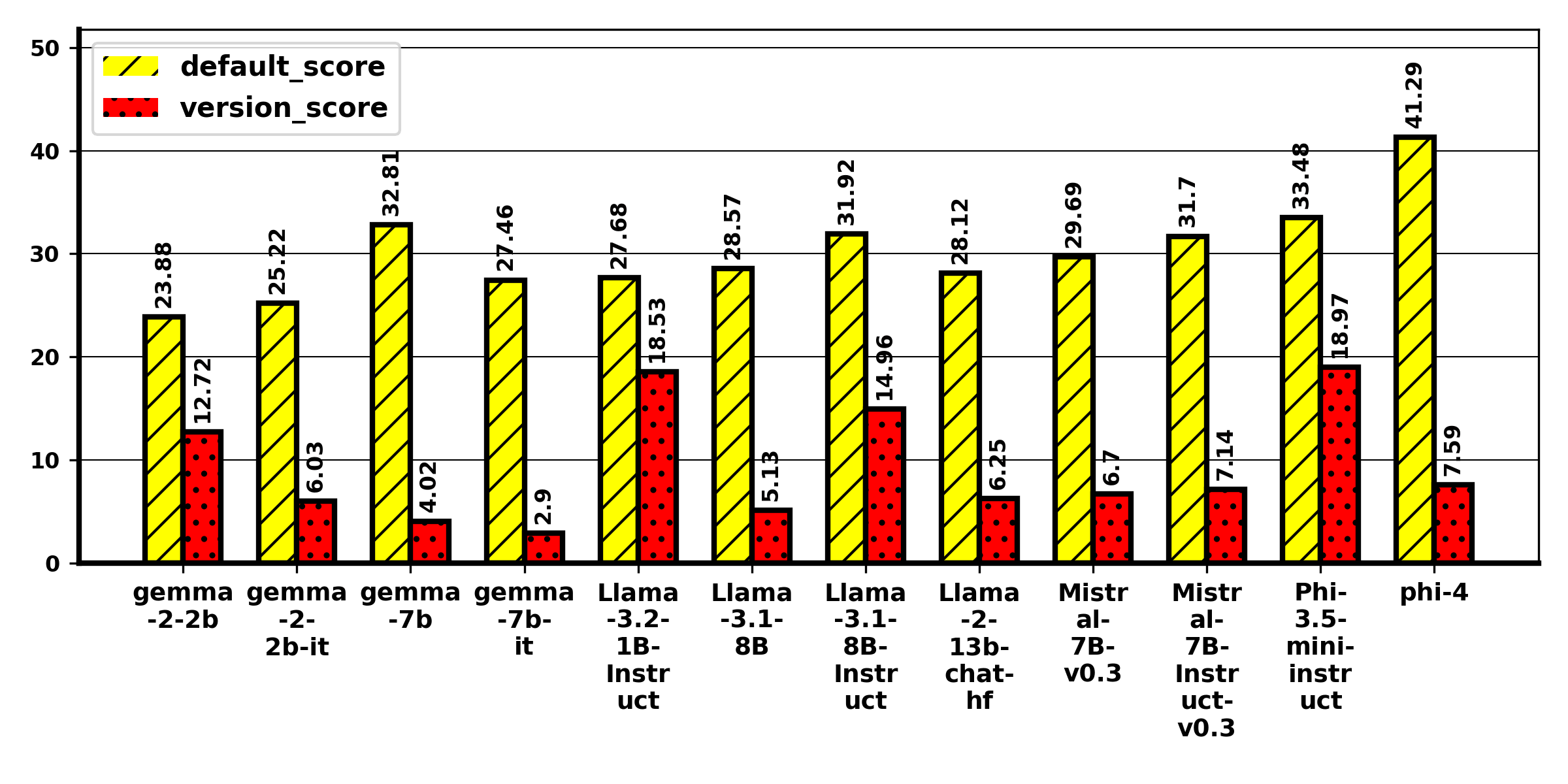}
    \vspace{-2em}
    \caption{$\mathbf{v_{none-correct}}$ results for GPQA benchmark.}
    \label{fig:results_vnonec_gpqa}
\end{minipage}
\hfill
\begin{minipage}[c]{0.48\linewidth}
\includegraphics[width=\columnwidth]{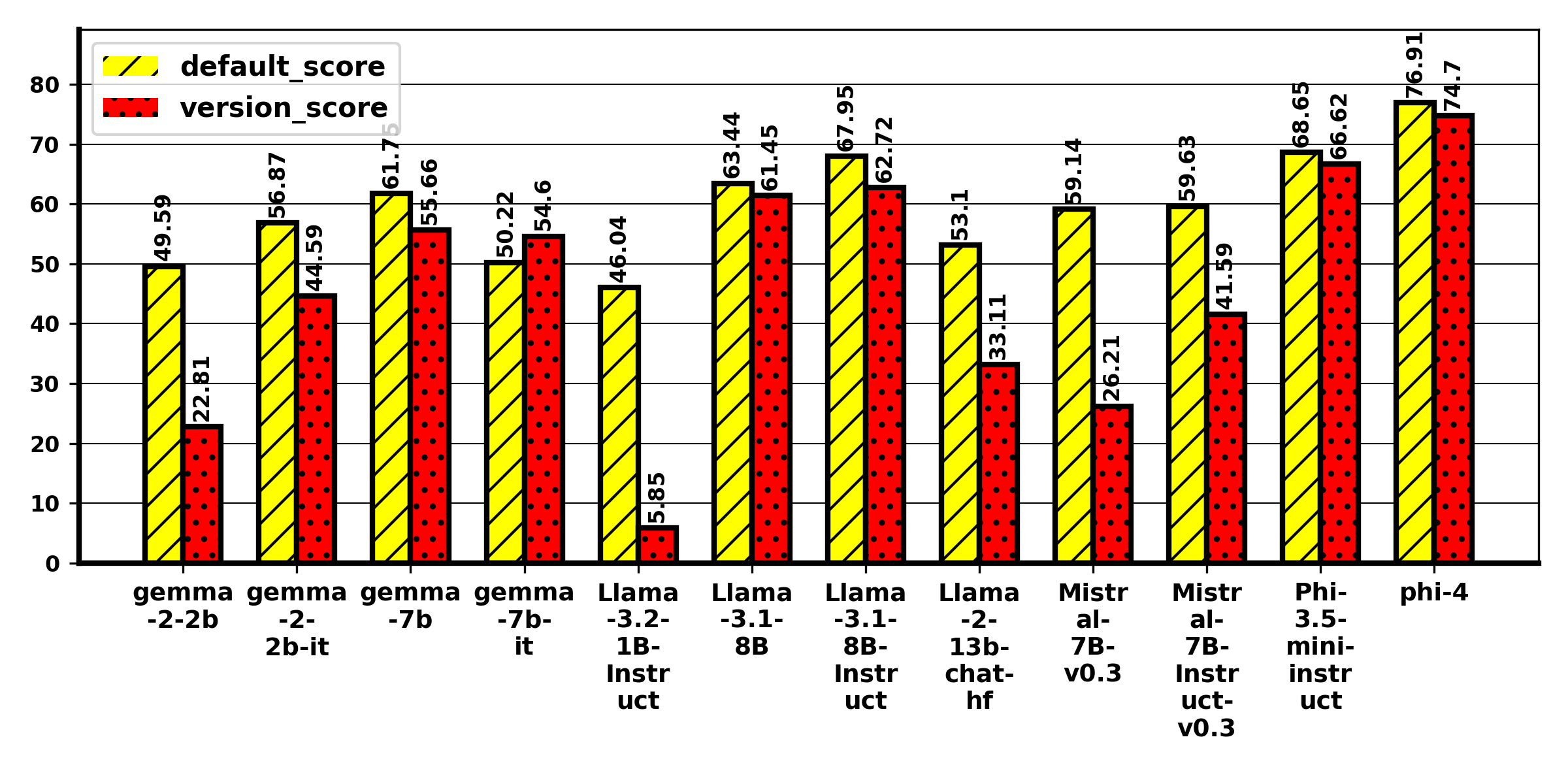}
   \vspace{-2em}
    \caption{$\mathbf{v_{all}}$ results for MMLU benchmark.}
    \label{fig:results_vall_mmlu}
\end{minipage}%
\end{figure*}

\subsection{Evaluation Setup and Infrastructure}  
We evaluate all original benchmarks and perturbed datasets using LLM-Eval-Harness~\cite{eval-harness}, modifying the configuration settings accordingly to accommodate different datasets. For the IfEval dataset, we implement a set of custom verification functions to handle various perturbations. We conduct experiments on either 4 NVIDIA A100 GPUs (40GB). The batch size varies based on the dataset. We adopt a zero-shot evaluation setup for GPQA, MMLU, MUSR, and IfEval, while MMLU-Pro is evaluated using a few-shot setup with $n = 5$, where $n$ denotes the number of few-shot examples. 

\begin{figure*}[ht]
\begin{minipage}[c]{0.48\linewidth}
\includegraphics[width=\columnwidth]{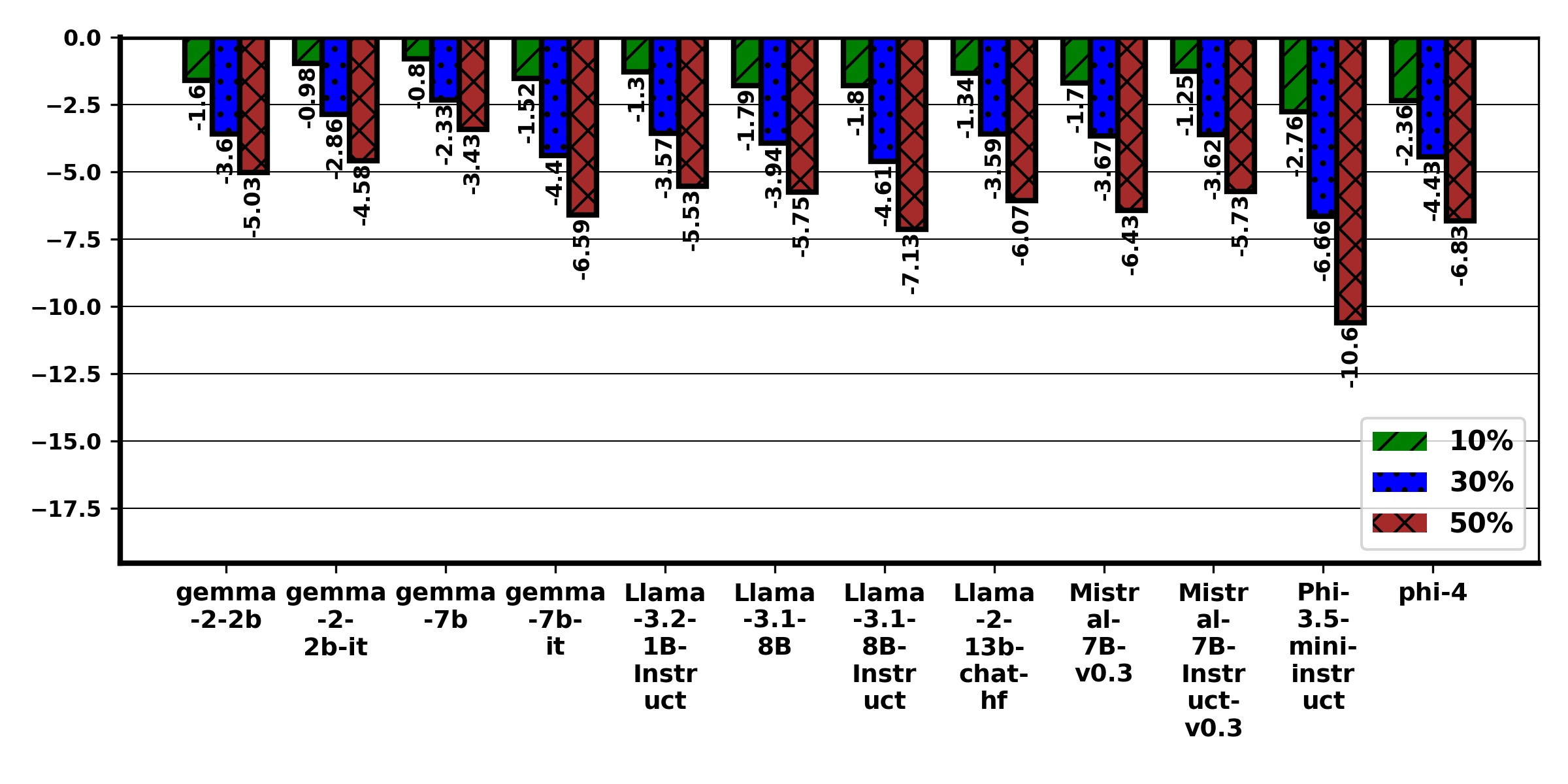}
    \vspace{-2em}
    \caption{$\mathbf{v_0}$ (Typographical errors) results on MMLU.}
    \label{fig:results_v0_mmlu}
\end{minipage}
\hfill
\begin{minipage}[c]{0.48\linewidth}
\includegraphics[width=\columnwidth]{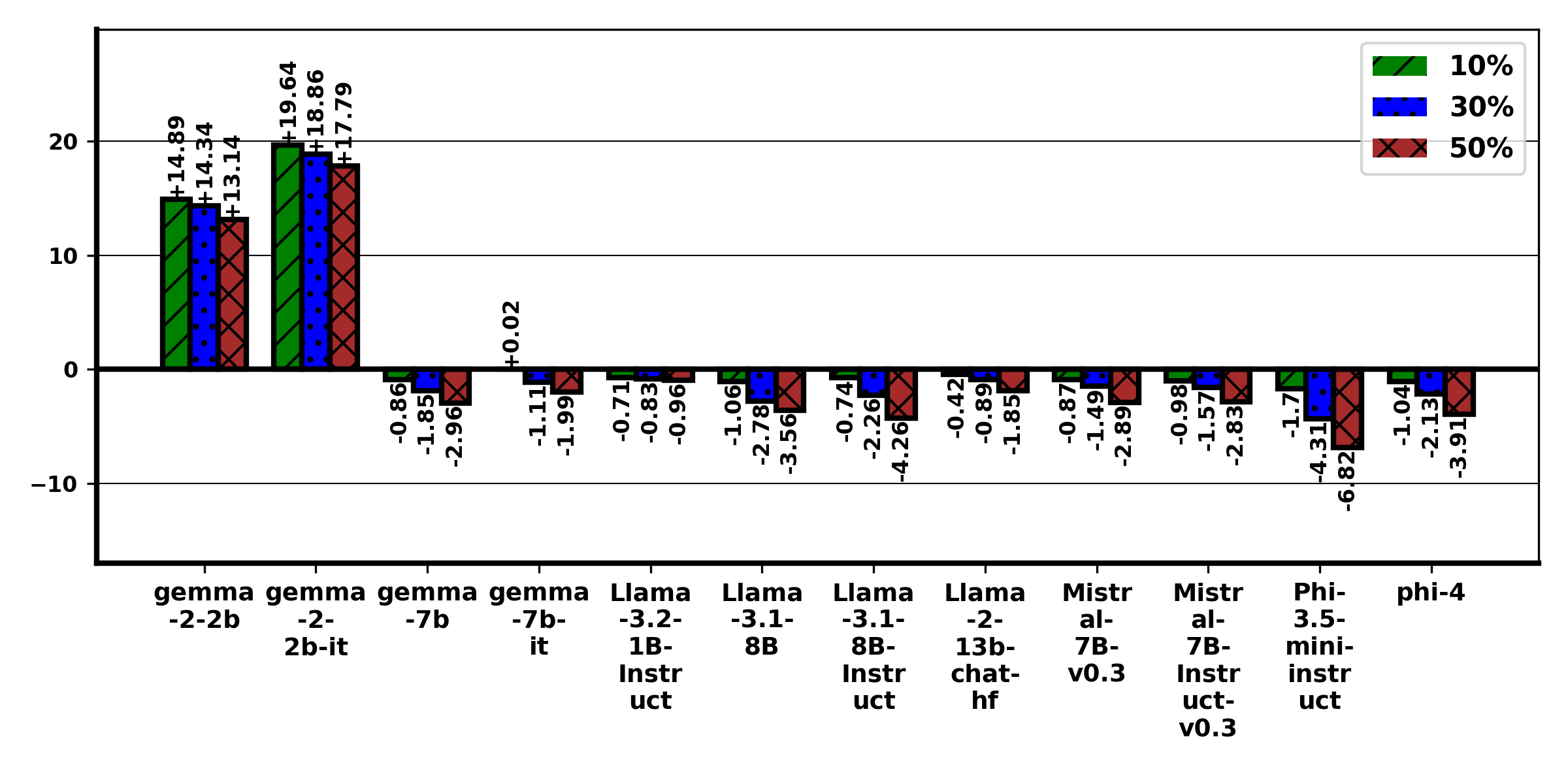}
   \vspace{-2em}
    \caption{$\mathbf{v_2}$ (Letter Removal) results on MMLU Pro.}
    \label{fig:results_v2_mmlu_pro}
\end{minipage}%
\end{figure*}

\subsection{Results} 
We first evaluate all selected models on the original benchmark datasets without perturbations to establish baseline scores. Next, we assess their performance on the perturbed datasets. The results are organized into two sections: structured answer selection and open-ended generation. 
\subsubsection{Structured answer selection}
\noindent\textbf{Impact of Question Perturbations} Typographical errors $\mathbf{v_0}$ significantly impact model performance, particularly for MMLU, where accuracy decreases consistently with higher perturbation rates. Phi-3.5-mini-Instruct is the most affected, experiencing a 10.6\% drop in accuracy (refer Figure~\ref{fig:results_v0_mmlu}), likely due to its reliance on synthetic datasets and filtered web content, which may not expose it to noisy text variations. A similar decline is observed for LLaMA-3.1-8B-Instruct, reinforcing the idea that robustness to typographical errors is not necessarily correlated with model size. In contrast, for MMLU-Pro, Gemma-2-2B and Gemma-2-2B-it exhibit better robustness and even surpass their baseline scores. For GPQA, no clear trend emerges, with minor fluctuations in accuracy indicating that the dataset's inherently complex, domain-specific questions overshadow the impact of typographical errors. In MUSR, Phi-4 exhibits the largest performance drop (9.79\%), while smaller models like LLaMA-3.2-1B-Instruct and Mistral-7B-it show slight accuracy improvements, hinting at potential differences in how models handle natural language reasoning under noisy conditions.  

With Letter duplication $\mathbf{v_1}$, performance degrades with increasing perturbation rates, but with some notable differences. In MMLU, larger models of the same version tend to be slightly more robust than their smaller counterparts; for example, Gemma-2-7B outperforms Gemma-2-2B. It suggests that while larger models do not always improve robustness, they may exhibit slight advantages in handling minor character-level noise. In MMLU-Pro, Gemma-2-2B and Gemma-2-2B-it remain the most robust models, maintaining performance close to the baseline even with 50\% perturbation. Unlike typographical errors, letter duplication does not cause significant accuracy changes in GPQA, likely because of the question complexity and domain knowledge requirements, that dominate model performance. In MUSR, Phi-4 again experiences the largest drop (6.88\%), while Gemma-7B-it and Mistral-7B-it show slight gains, suggesting that certain instruction-tuned models may better adapt to redundant text patterns.  

As with previous perturbations, for $\mathbf{v_2}$ accuracy declines across all models in MMLU, with Phi-3.5-mini-Instruct showing the steepest drop (8.85\%). The trend in MMLU-Pro remain consistent with other perturbations, where Gemma-2-2B and Gemma-2-2B-it are more resilient as shown in Figure~\ref{fig:results_v2_mmlu_pro}. GPQA exhibits no clear robustness pattern, further reinforcing the idea that its complexity overshadows minor text perturbations. For MUSR, Phi-4 shows the largest decline (7.28\%), while no strong patterns emerge across other models.  

We find that robustness to perturbations is model and dataset dependent. Some models, such as Gemma-2-2B, exhibit resilience across multiple benchmarks. Others, like Phi-3.5-mini-Instruct and Phi-4, are more vulnerable to noise, due to dependency in synthetic data training. Furthermore, the lack of a clear correlation between model size and robustness suggests that pretraining data quality play a crucial role in handling perturbations. 

\noindent\textbf{Impact of Choices Perturbations} Across all benchmarks except MUSR, replacing an incorrect choice with "None of these" $\mathbf{v_{none}}$ leads to a slight improvement in accuracy for most models. The models might have a bias toward recognizing "None of these" as a non-informative choice, allowing them to focus on the remaining options. However, for the MUSR benchmark, the perturbation results in a decline in accuracy.

When an incorrect option is replaced with "All of the above" $\mathbf{v_{all}}$, a general decline in accuracy is observed for MMLU and MUSR benchmarks as shown in Figure~\ref{fig:results_vall_mmlu}. This drop could stem from the fact that "All of the above" introduces an implicit relationship between choices, which may not have existed in the original question format, confusing the model’s decision process. In contrast, for the MMLU-Pro benchmark, there are only slight deviations from the default scores, with some models showing a minor increase or decrease in accuracy. The GPQA benchmark does not exhibit a clear trend, though the gemma-2-2b model shows a significant accuracy drop of approximately 12\%, suggesting that certain models may struggle with reinterpreting the answer structure. Replacing a correct option with "None of these" $\mathbf{v_{none-correct}}$ has the most significant negative impact on performance across all benchmarks and models, leading to a substantial decline in accuracy as shown in Figure~\ref{fig:results_vnonec_gpqa}. This result is expected, as removing a correct choice and replacing it with an ambiguous alternative forces the model to discard the right answer, leading to increased misclassification. Interestingly, the MUSR benchmark shows the opposite trend, with accuracy increasing. Finally, perturbing the choices by introducing variations of "Both A and B" $\mathbf{v_{both}}$ generally results in a decline in accuracy for MMLU and MUSR benchmarks, indicating that models struggle with interpreting multiple-choice dependencies. In MMLU-Pro, the accuracy deviations remain minor, suggesting that the format of this dataset is less sensitive to such changes. For GPQA, most models show a decline, except for Phi models and gemma-7b-it, where scores increase slightly. The largest drop is observed for the gemma-2-2b model, with a 9.71\% decrease in accuracy, reinforcing the idea that smaller models are more sensitive to perturbations that alter the logical structure of answer choices.  It is observed that models are biased towards "None of These" and "All of These" when number of answer choices are either 2 or 3, irrespective of the remaining options, showing unlikely behaviour in MUSR scores.  

The results highlight that while LLMs demonstrate some robustness to superficial changes in answer formats, modifications that disrupt the logical structure of choices or remove correct options significantly degrade performance. 
 
\begin{figure}[b]
\begin{minipage}[c]{0.48\linewidth}
\includegraphics[width=\columnwidth]{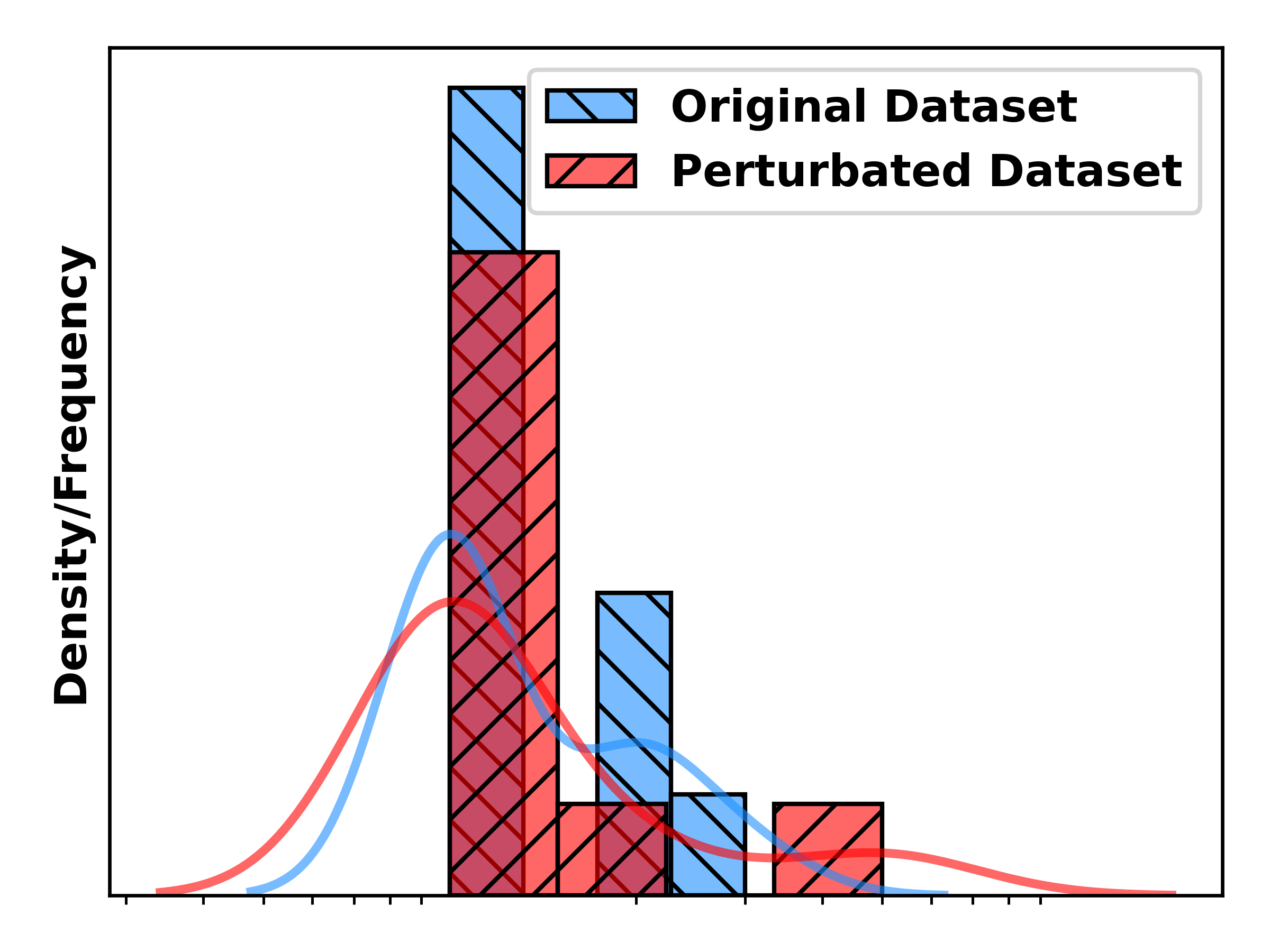}
\label{fig:KS_test}
\end{minipage}
\hfill
\begin{minipage}[c]{0.48\linewidth}
\includegraphics[width=\columnwidth]{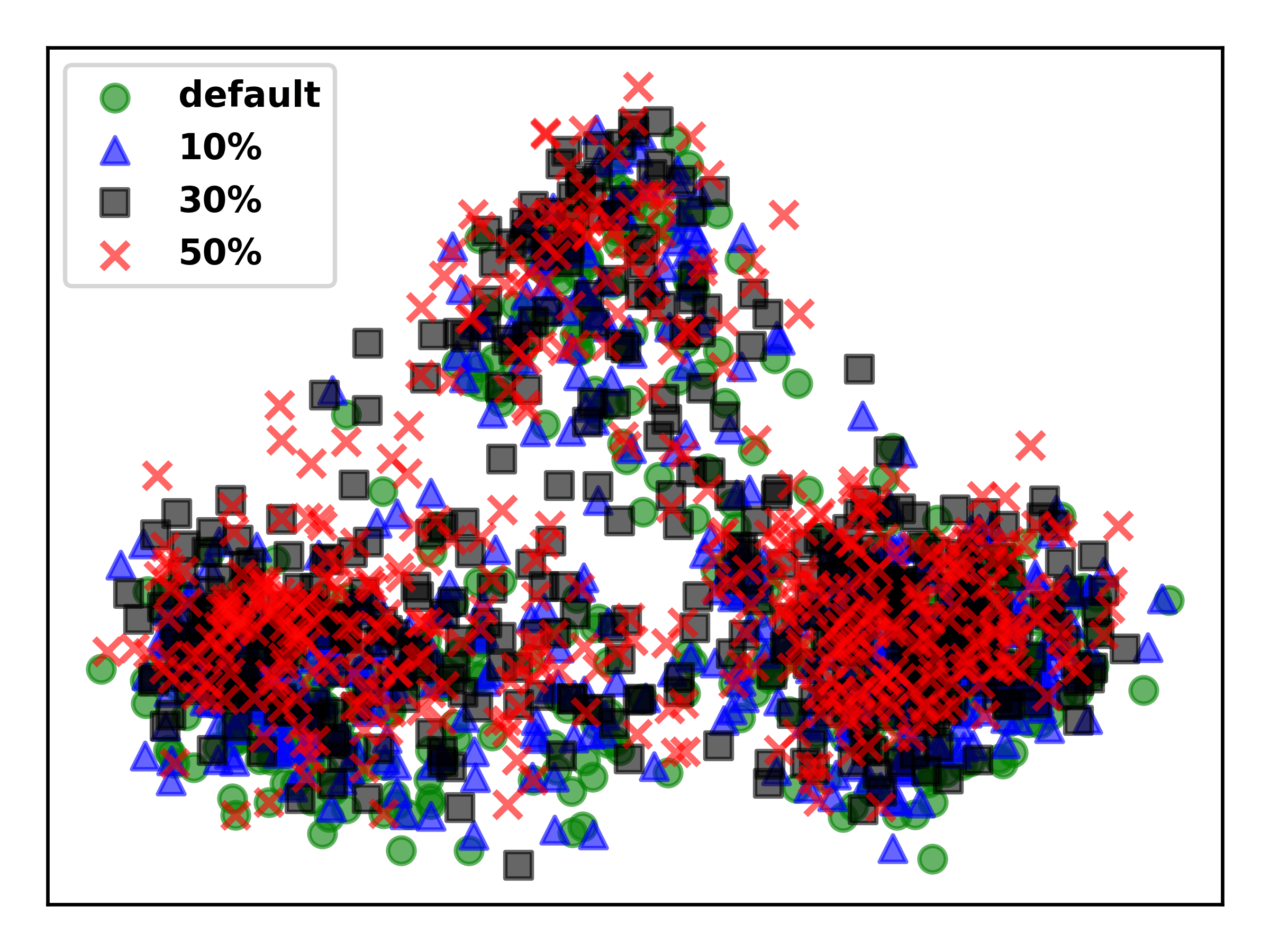}
\label{fig:pcs_embedding}
\end{minipage}%
\vspace{-2em}
\caption{Distribution of original and perturbed data}
\label{fig:data_distribution}
\end{figure}

\subsubsection{Open-ended generation}
We evaluate the impact of perturbations on open-ended generation using the IfEVal dataset and observe degradation of 2.85\%. in performance across models. The extent of degradation depends on the model type, with instruct-tuned models exhibiting higher robustness compared to base models. In particular, instruct models consistently outperform their base counterparts in accuracy. The trend is evident even when comparing models of different scales. For instance, Phi-4, despite being a relatively large model with 14B parameters, demonstrates a notably low performance, suggesting that scale alone does not ensure robustness to perturbations. Instead, alignment and fine-tuning strategies play a crucial role in maintaining accuracy under distribution shifts. The effect of perturbations on base models is highly variable. Among different model families, we observe that LLaMA-based models exhibit the largest fluctuations, with an average accuracy drop of approximately 4\%. This suggests that these models are more sensitive to input modifications, possibly due to the nature of their pretraining or the lack of strong instruction tuning. The observed variability further highlights the limitations of base models in handling adversarial or altered inputs, underscoring the importance of robustness-focused training strategies.  

We find that while perturbations introduce performance degradation across models, instruct-tuned models are more resilient, whereas base models, particularly those in the LLaMA family, exhibit significant variability under perturbations. We provide the detailed results and plots in appendix. 

\section{Discussion and Analysis}

We apply the two-sample non-parametric K-S test to determine whether the original and perturbed datasets follow the same distribution. Specifically, for the MMLU-Pro dataset, we consider the questions from each sample and their perturbed counterparts. We preprocess the text by removing special characters and stopwords, followed by lemmatization. We then compute the word frequency distributions and perform the K-S test. The K-S test yields a p-value of 0.9750, indicating that the original and perturbed datasets follow the same underlying distribution, as the high p-value ($\geq 0.05$) suggests no significant statistical difference between them. The resulting distributional differences (log scaled) between the original and perturbed datasets are visualized in Figure~\ref{fig:data_distribution}.  

Additionally, we analyze the semantic similarity between original and perturbed questions by computing their embeddings. Figure~\ref{fig:data_distribution} presents the projected embeddings of original and perturbed questions after applying principal component analysis. The minimal divergence between the two distributions indicates that the perturbations introduce controlled modifications while preserving the overall statistical properties of the original dataset.

To ensure completeness and assess variability in the results, we generate datasets using three different random seeds. For instance, in the case of the MMLU dataset, we obtain average standard deviations of 0.47, 0.48, and 1.01. This consistency in observations across multiple perturbation runs indicates the stability of our findings.

\section{Conclusion}
In this work, we provide a comprehensive evaluation of LLM robustness by analyzing their sensitivity to perturbations in multiple-choice question datasets and instruction-following tasks. Through systematic perturbations including typographical errors, choice modifications, and instruction variability, we observe that LLMs exhibit varying degrees of robustness depending on the nature of the perturbation and the benchmark used. While some perturbations, such as minor typographical errors, lead to moderate accuracy drops, others, such as replacing correct choices with None of these in MCQs, cause significant performance degradation. Interestingly, larger models do not consistently demonstrate higher robustness, suggesting that scale alone does not confer stability under perturbed conditions. The findings underscore the need for improved training strategies and evaluation methodologies that account for real-world input variability. In Future work, we intend to explore the adaptive fine-tuning techniques and adversarial training strategies to enhance LLM resilience against structured perturbations.

\section*{Limitations}
While our study evaluates the robustness of models against perturbations, several aspects remain unexplored. One key limitation is that our analysis does not consider perturbations involving synonym and antonym substitutions or broader semantic shifts in word distributions. Prior work has investigated the impact of such modifications on model robustness, highlighting their significance in evaluating language understanding; however, we leave this aspect for future research.  

Additionally, our evaluation does not extend to larger models beyond 14B size. Given that larger-scale models often exhibit different generalization behaviors, assessing whether increased parameter counts improve robustness against perturbations remains an open question.  

Finally, while we analyze variability in open-ended generation, our study is limited in the number of samples considered. A more extensive analysis with a larger sample set would provide deeper insights into the consistency and reliability of model outputs under perturbations. Future work should address these limitations by incorporating a broader range of perturbation types, evaluating larger models, and expanding the dataset size to capture more comprehensive trends.

\bibliography{custom}

\appendix

\section{Example Appendix}
\label{sec:appendix}

\subsection{IfEval samples}
Below is an example input to the model, taken from their paper.
\fbox{
    \begin{minipage}{\linewidth}
        \textbf{Prompt} : Who built the first artificial ice rink? Please include the keys $\left(1\right)$ Name $\left(2\right)$ Location
        and $\left(3\right)$ Year. \textit{Use less than 487 words.} 
    \end{minipage} 
}
\captionsetup{type=figure} 
\captionof{figure}{The fig shows a prompt with, verifiable instructions in italic.}
We encourage you to read the IfEval paper for more clarity on the dataset.

\fbox{
    \begin{minipage}{\linewidth}
        \textbf{Sample 1}:\\
        \textbf{prompt}: I am planning a trip to Japan, and I would like thee to write an itinerary for my journey in a Shakespearean style. You are \textbf{not} allowed to use any \textcolor{red}{commas} in your response.\\
        \textbf{instruction\_id\_list}: $\left[ ``punctuation:\textcolor{red}{no\_comma}" \right]$\\
        \textbf{changed\_prompt}: I am planning a trip to Japan, and I would like thee to write an itinerary for my journey in a Shakespearean style. You are \textbf{not} allowed to use any \textcolor{green}{semicolons} in your response.\\
        \textbf{changed\_instruction\_id\_list}: $\left[ ``punctuation:\textcolor{green}{no\_semicolon}" \right]$\\
        
        \textbf{Sample 2}:\\
        \textbf{prompt}: Given the sentence "Two young boys with toy guns and horns." can you ask a question? Please ensure that your response is in English, and in all lowercase letters. \textbf{No} \textcolor{red}{capital} letters are allowed.\\
        \textbf{instruction\_id\_list}: $\left[ ``change\_case:\textcolor{red}{english\_lowercase}" \right]$\\
        \textbf{changed\_prompt}: Given the sentence "Two young boys with toy guns and horns." can you ask a question? Please ensure that your response is in English, and in all lowercase letters. \textbf{No} \textcolor{green}{lowercase} letters are allowed.\\
        \textbf{changed\_instruction\_id\_list}: $\left[ ``change\_case:\textcolor{green}{english\_uppercase}" \right]$
    \end{minipage} 
}

\captionsetup{type=figure} 
\captionof{figure}{The fig shows a few samples with changes we introduce}

\vspace{0.15cm}

\subsection{Summarized results for Modified Questions}

\begin{table*}[]
    \centering
    \resizebox{\textwidth}{!}{
        \begin{tabular}{c|c|ccc|ccc|ccc}
            \hline
            \multirow{2}{*}{\textbf{Models}}   & \multirow{2}{*}{\textbf{Default}} & \multicolumn{3}{c|}{\textbf{$\mathbf{v_0}$}}              & \multicolumn{3}{c|}{\textbf{$\mathbf{v_1}$}}              & \multicolumn{3}{c}{\textbf{$\mathbf{v_2}$}}               \\ \cline{3-11} 
                                               &                                   & \textbf{10\%} & \textbf{30\%} & \textbf{50\%} & \textbf{10\%} & \textbf{30\%} & \textbf{50\%} & \textbf{10\%} & \textbf{30\%} & \textbf{50\%} \\ \hline
            google/gemma-2-2b                  & 23.88                             & 24.55         & 22.77         & 23.88         & 22.77         & 20.76         & 23.88         & 23.66         & 23.21         & 25.45         \\
            google/gemma-2-2b-it               & 25.22                             & 27.90         & 27.46         & 27.01         & 29.46         & 29.02         & 27.23         & 28.12         & 29.24         & 27.46         \\
            google/gemma-7b                    & 32.81                             & 29.69         & 31.03         & 31.03         & 31.92         & 29.24         & 31.92         & 32.81         & 31.92         & 33.04         \\
            google/gemma-7b-it                 & 27.46                             & 29.24         & 29.46         & 28.57         & 29.69         & 29.02         & 29.02         & 27.23         & 29.69         & 30.36         \\ \hline
            meta-llama/Llama-3.2-1B-Instruct   & 27.68                             & 27.68         & 29.46         & 27.46         & 26.56         & 27.90         & 25.45         & 27.01         & 27.90         & 27.01         \\
            meta-llama/Llama-3.1-8B            & 28.57                             & 28.35         & 28.12         & 30.80         & 28.35         & 28.12         & 29.02         & 28.79         & 31.25         & 31.03         \\
            meta-llama/Llama-3.1-8B-Instruct   & 34.38                             & 30.80         & 29.69         & 32.81         & 29.46         & 32.14         & 30.80         & 29.91         & 32.59         & 30.13         \\
            meta-llama/Llama-2-13b-chat-hf     & 28.12                             & 27.68         & 28.35         & 27.90         & 27.90         & 27.46         & 27.23         & 28.35         & 29.24         & 29.69         \\ \hline
            mistralai/Mistral-7B-v0.3          & 29.69                             & 29.46         & 29.91         & 30.58         & 30.58         & 30.36         & 30.36         & 30.80         & 30.13         & 29.46         \\
            mistralai/Mistral-7B-Instruct-v0.3 & 31.70                             & 32.14         & 30.36         & 29.69         & 32.14         & 33.04         & 31.47         & 31.03         & 31.70         & 30.13         \\ \hline
            microsoft/phi-4                    & 41.29                             & 38.39         & 37.95         & 38.84         & 38.84         & 38.17         & 38.39         & 39.73         & 39.51         & 36.83         \\
            microsoft/Phi-3.5-mini-instruct    & 33.48                             & 29.91         & 31.25         & 28.35         & 31.47         & 31.25         & 29.69         & 31.47         & 31.47         & 31.70         \\ \hline
            \end{tabular}
    }
    \caption{Scores for Structured Answer Selection with modified GPQA Questions}
\end{table*}

\begin{table*}[]
    \centering
    \resizebox{\textwidth}{!}{
        \begin{tabular}{c|c|ccc|ccc|ccc}
        \hline
        \multirow{2}{*}{\textbf{Models}}   & \multirow{2}{*}{\textbf{Default}} & \multicolumn{3}{c|}{\textbf{$\mathbf{v_0}$}}              & \multicolumn{3}{c|}{\textbf{$\mathbf{v_1}$}}              & \multicolumn{3}{c}{\textbf{$\mathbf{v_2}$}}               \\ \cline{3-11} 
                                        &                                   & \textbf{10\%} & \textbf{30\%} & \textbf{50\%} & \textbf{10\%} & \textbf{30\%} & \textbf{50\%} & \textbf{10\%} & \textbf{30\%} & \textbf{50\%} \\ \hline
        google/gemma-2-2b                  & 7.35                              & 22.17         & 21.87         & 20.75         & 22.43         & 21.63         & 21.43         & 22.23         & 21.68         & 20.49         \\
        google/gemma-2-2b-it               & 9.38                              & 28.90         & 27.97         & 26.81         & 29.11         & 28.74         & 28.23         & 29.01         & 28.23         & 27.17         \\
        google/gemma-7b                    & 33                                & 32.10         & 31.10         & 29.85         & 32.52         & 31.76         & 31.52         & 32.13         & 31.14         & 30.04         \\
        google/gemma-7b-it                 & 25.19                             & 24.91         & 24            & 22.86         & 24.93         & 24.24         & 24.10         & 25.21         & 24.09         & 23.20         \\ \hline
        meta-llama/Llama-3.2-1B-Instruct   & 17.85                             & 17.84         & 17.05         & 16.69         & 18.10         & 17.40         & 17.11         & 17.15         & 17.02         & 16.89         \\
        meta-llama/Llama-3.1-8B            & 35.41                             & 34.56         & 32.88         & 31.58         & 34.72         & 33.82         & 32.73         & 34.35         & 32.62         & 31.85         \\
        meta-llama/Llama-3.1-8B-Instruct   & 40.92                             & 39.58         & 38.51         & 36.79         & 40.63         & 39.96         & 39.28         & 40.18         & 38.66         & 36.65         \\
        meta-llama/Llama-2-13b-chat-hf     & 25.94                             & 25.30         & 24.13         & 23.68         & 26.12         & 25.18         & 24.95         & 25.52         & 25.05         & 24.09         \\ \hline
        mistralai/Mistral-7B-v0.3          & 30.28                             & 30.02         & 28.10         & 27.42         & 29.75         & 29.30         & 28.83         & 29.40         & 28.79         & 27.39         \\
        mistralai/Mistral-7B-Instruct-v0.3 & 33.55                             & 32.91         & 31.23         & 29.96         & 32.62         & 33.04         & 33.05         & 32.57         & 31.98         & 30.72         \\ \hline
        microsoft/phi-4                    & 59.14                             & 57.69         & 56.78         & 54.27         & 58.76         & 58.21         & 57.24         & 58.10         & 57.01         & 55.24         \\
        microsoft/Phi-3.5-mini-instruct    & 44.31                             & 41.86         & 39.25         & 35.52         & 43.03         & 42.02         & 39.91         & 42.61         & 39.99         & 37.48         \\ \hline
        \end{tabular}
    }
    \caption{Scores for Structured Answer Selection with modified MMLU-PRO Questions}
\end{table*}

\begin{table*}[]
    \centering
    \resizebox{\textwidth}{!}{
        \begin{tabular}{c|c|ccc|ccc|ccc}
            \hline
            \multirow{2}{*}{\textbf{Models}}   & \multirow{2}{*}{\textbf{Default}} & \multicolumn{3}{c|}{\textbf{$\mathbf{v_0}$}}                 & \multicolumn{3}{c|}{\textbf{$\mathbf{v_1}$}}                 & \multicolumn{3}{c}{\textbf{$\mathbf{v_2}$}}                  \\ \cline{3-11} 
                                               &                                   & \textbf{10.00} & \textbf{30.00} & \textbf{50.00} & \textbf{10.00} & \textbf{30.00} & \textbf{50.00} & \textbf{10.00} & \textbf{30.00} & \textbf{50.00} \\ \hline
            google/gemma-2-2b                  & 49.59                             & 47.98          & 45.98          & 44.56          & 48.44          & 47.54          & 46.56          & 48.03          & 46.27          & 44.77          \\
            google/gemma-2-2b-it               & 56.87                             & 55.90          & 54.01          & 52.29          & 56.56          & 55.48          & 54.66          & 56.33          & 54.54          & 52.35          \\
            google/gemma-7b                    & 61.75                             & 60.95          & 59.42          & 58.33          & 61.48          & 60.63          & 60.11          & 61.17          & 59.83          & 57.95          \\
            google/gemma-7b-it                 & 50.22                             & 48.70          & 45.82          & 43.63          & 49.07          & 48.01          & 46.72          & 48.79          & 46.70          & 44.54          \\ \hline
            meta-llama/Llama-3.2-1B-Instruct   & 46.04                             & 44.74          & 42.47          & 40.51          & 44.94          & 43.93          & 42.76          & 44.56          & 43.04          & 41.47          \\
            meta-llama/Llama-3.1-8B            & 63.44                             & 61.64          & 59.50          & 57.69          & 62.88          & 61.85          & 60.87          & 62.01          & 60.03          & 58.14          \\
            meta-llama/Llama-3.1-8B-Instruct   & 68.02                             & 66.22          & 63.41          & 60.89          & 67.40          & 65.92          & 65.16          & 66.37          & 64.53          & 61.99          \\
            meta-llama/Llama-2-13b-chat-hf     & 53.10                             & 51.77          & 49.52          & 47.03          & 52.40          & 51.51          & 50.61          & 51.79          & 50.13          & 48.73          \\ \hline
            mistralai/Mistral-7B-v0.3          & 59.14                             & 57.44          & 55.47          & 52.71          & 58.49          & 57.90          & 57.02          & 58.21          & 56.44          & 53.64          \\
            mistralai/Mistral-7B-Instruct-v0.3 & 59.63                             & 58.37          & 56             & 53.90          & 59.26          & 58.67          & 57.88          & 58.99          & 56.95          & 55.05          \\ \hline
            microsoft/phi-4                    & 76.91                             & 74.55          & 72.48          & 70.08          & 76.02          & 75.37          & 74.53          & 74.86          & 72.96          & 70.54          \\
            microsoft/Phi-3.5-mini-instruct    & 68.65                             & 65.89          & 61.99          & 58.05          & 67.34          & 65.74          & 63.59          & 66.79          & 63.18          & 59.80          \\ \hline
            \end{tabular}
    }
    \caption{Scores for Structured Answer Selection with modified MMLU Questions}
\end{table*}

\begin{table*}[]
    \centering
    \resizebox{\textwidth}{!}{
        \begin{tabular}{c|c|ccc|ccc|ccc}
            \hline
            \multirow{2}{*}{\textbf{Models}}   & \multirow{2}{*}{\textbf{Default}} & \multicolumn{3}{c|}{\textbf{$\mathbf{v_0}$}}                 & \multicolumn{3}{c|}{\textbf{$\mathbf{v_1}$}}                 & \multicolumn{3}{c}{\textbf{$\mathbf{v_2}$}}                  \\ \cline{3-11} 
                                               &                                   & \textbf{10.00} & \textbf{30.00} & \textbf{50.00} & \textbf{10.00} & \textbf{30.00} & \textbf{50.00} & \textbf{10.00} & \textbf{30.00} & \textbf{50.00} \\ \hline
            google/gemma-2-2b                  & 42.59                             & 40.21          & 39.29          & 38.49          & 40.21          & 40.08          & 40.48          & 40.34          & 39.95          & 40.08          \\
            google/gemma-2-2b-it               & 42.99                             & 42.33          & 42.06          & 41.67          & 42.72          & 40.87          & 41.93          & 41.01          & 42.99          & 43.25          \\
            google/gemma-7b                    & 40.21                             & 39.15          & 37.17          & 37.17          & 37.57          & 39.15          & 37.96          & 37.96          & 35.45          & 36.51          \\
            google/gemma-7b-it                 & 44.71                             & 45.90          & 44.58          & 44.71          & 44.97          & 46.03          & 45.37          & 44.58          & 45.63          & 43.78          \\ \hline
            meta-llama/Llama-3.2-1B-Instruct   & 34.39                             & 35.58          & 35.32          & 35.71          & 34.92          & 34.52          & 34.26          & 34.39          & 32.80          & 34.39          \\
            meta-llama/Llama-3.1-8B            & 38.62                             & 38.23          & 36.24          & 37.17          & 38.23          & 36.77          & 37.43          & 37.17          & 37.17          & 36.64          \\
            Meta-Llama-3.1-8B-Instruct         & 40.08                             & 38.23          & 36.90          & 36.38          & 38.49          & 38.49          & 39.42          & 39.15          & 36.24          & 35.45          \\
            meta-llama/Llama-2-13b-chat-hf     & 41.14                             & 41.27          & 41.67          & 40.34          & 40.21          & 40.74          & 40.21          & 40.48          & 41.01          & 39.81          \\ \hline
            mistralai/Mistral-7B-v0.3          & 40.34                             & 38.36          & 37.96          & 36.77          & 39.29          & 38.76          & 38.36          & 39.81          & 38.10          & 38.76          \\
            mistralai/Mistral-7B-Instruct-v0.3 & 47.22                             & 48.41          & 47.62          & 47.75          & 48.81          & 48.54          & 48.28          & 47.09          & 48.68          & 46.83          \\ \hline
            microsoft/phi-4                    & 50.13                             & 44.84          & 41.80          & 40.34          & 44.58          & 44.18          & 43.25          & 45.90          & 42.86          & 42.86          \\
            microsoft/Phi-3.5-mini-instruct    & 43.52                             & 41.93          & 42.86          & 42.72          & 42.06          & 42.86          & 42.46          & 42.72          & 42.86          & 44.05          \\ \hline
            \end{tabular}
    }
    \caption{Scores for Structured Answer Selection with modified MUSR Narrations}
\end{table*}

\subsection{Summarized results for Modified Choices}

\begin{table}[]
    \centering
    \resizebox{\columnwidth}{!}{
        \begin{tabular}{c|c|c|c|c|c}
            \hline
            \textbf{Models}                    & \textbf{Default} & \textbf{$\mathbf{v_{none}}$} & \textbf{$\mathbf{v_{all}}$} & \textbf{$\mathbf{v_{none-correct}}$} & \textbf{$\mathbf{v_{both}}$} \\ \hline
            google/gemma-2-2b                  & 23.88            & 28.79       & 12.72       & 12.72       & 14.73       \\
            google/gemma-2-2b-it               & 25.22            & 30.36       & 29.46       & 6.03        & 24.55       \\
            google/gemma-7b                    & 32.81            & 39.96       & 35.71       & 4.02        & 30.36       \\
            google/gemma-7b-it                 & 27.46            & 36.83       & 34.60       & 2.90        & 30.36       \\ \hline
            meta-llama/Llama-3.2-1B-Instruct   & 27.68            & 30.58       & 16.74       & 18.53       & 19.87       \\
            meta-llama/Llama-3.1-8B            & 28.57            & 37.72       & 22.99       & 5.13        & 26.12       \\
            meta-llama/Llama-3.1-8B-Instruct   & 31.92            & 32.59       & 28.12       & 14.96       & 29.91       \\
            meta-llama/Llama-2-13b-chat-hf     & 28.12            & 37.28       & 25.22       & 6.25        & 20.54       \\ \hline
            mistralai/Mistral-7B-v0.3          & 29.69            & 35.27       & 25.67       & 6.70        & 24.33       \\
            mistralai/Mistral-7B-Instruct-v0.3 & 31.70            & 33.93       & 34.60       & 7.14        & 31.03       \\ \hline
            microsoft/phi-4                    & 41.29            & 44.87       & 44.64       & 7.59        & 42.86       \\
            microsoft/Phi-3.5-mini-instruct    & 33.48            & 35.94       & 37.28       & 18.97       & 34.82       \\ \hline      
            \end{tabular}
    }
    \caption{Scores for Structured Answer Selection with modified GPQA Choices}
\end{table}

\begin{table}[]
    \centering
    \resizebox{\columnwidth}{!}{
        \begin{tabular}{c|c|c|c|c|c}
            \hline
            \textbf{models}                    & \textbf{Default} & \textbf{$\mathbf{v_{none}}$} & \textbf{$\mathbf{v_{all}}$} & \textbf{$\mathbf{v_{none-correct}}$} & \textbf{$\mathbf{v_{both}}$} \\ \hline
            google/gemma-2-2b                  & 42.59            & 25.93       & 21.69       & 55.82       & 38.10       \\
            google/gemma-2-2b-it               & 42.99            & 26.06       & 22.75       & 62.96       & 37.30       \\
            google/gemma-7b                    & 40.21            & 37.04       & 16.01       & 36.38       & 33.07       \\
            google/gemma-7b-it                 & 44.71            & 29.63       & 22.35       & 53.70       & 39.42       \\ \hline
            meta-llama/Llama-3.2-1B-Instruct   & 34.39            & 13.76       & 9.92        & 66.67       & 29.89       \\
            meta-llama/Llama-3.1-8B            & 38.62            & 20.77       & 16.53       & 63.62       & 31.88       \\
            meta-llama/Llama-3.1-8B-Instruct   & 40.08            & 19.18       & 17.72       & 66.27       & 33.20       \\
            meta-llama/Llama-2-13b-chat-hf     & 41.14            & 25.79       & 19.58       & 60.45       & 35.71       \\ \hline
            mistralai/Mistral-7B-v0.3          & 40.34            & 22.09       & 20.11       & 61.77       & 35.98       \\
            mistralai/Mistral-7B-Instruct-v0.3 & 47.22            & 48.54       & 24.74       & 41.01       & 40.61       \\ \hline
            microsoft/phi-4                    & 50.13            & 55.95       & 41.67       & 31.08       & 43.12       \\
            microsoft/Phi-3.5-mini-instruct    & 43.52            & 36.11       & 16.67       & 40.08       & 36.51       \\ \hline
            \end{tabular}
    }
    \caption{Scores for Structured Answer Selection with modified MUSR Choices}
\end{table}

\begin{table}[]
    \centering
    \resizebox{\columnwidth}{!}{
        \begin{tabular}{c|c|c|c|c|c}
            \hline
            \textbf{models}                    & \textbf{Default} & \textbf{$\mathbf{v_{none}}$} & \textbf{$\mathbf{v_{all}}$} & \textbf{$\mathbf{v_{none-correct}}$} & \textbf{$\mathbf{v_{both}}$} \\ \hline
            google/gemma-2-2b                  & 7.35             & 7.88        & 7.31        & 0.99        & 7.18        \\
            google/gemma-2-2b-it               & 9.38             & 9.96        & 9.53        & 2.24        & 8.72        \\
            google/gemma-7b                    & 33               & 33.97       & 32.75       & 5.43        & 33.10       \\
            google/gemma-7b-it                 & 25.19            & 27.49       & 27.69       & 0.97        & 21.18       \\ \hline
            meta-llama/Llama-3.2-1B-Instruct   & 17.85            & 19.46       & 16.17       & 2.14        & 16.52       \\
            meta-llama/Llama-3.1-8B            & 35.41            & 35.51       & 33.37       & 9.64        & 33.21       \\
            meta-llama/Llama-3.1-8B-Instruct   & 40.82            & 41.44       & 41.03       & 11.34       & 41.15       \\
            meta-llama/Llama-2-13b-chat-hf     & 25.94            & 27.91       & 26.35       & 2.85        & 26.15       \\ \hline
            mistralai/Mistral-7B-v0.3          & 30.28            & 31.45       & 30.20       & 2.86        & 31.33       \\
            mistralai/Mistral-7B-Instruct-v0.3 & 33.55            & 34.92       & 34.60       & 4.55        & 34.32       \\ \hline
            microsoft/phi-4                    & 59.14            & 58.69       & 59.23       & 20.50       & 58.94       \\
            microsoft/Phi-3.5-mini-instruct    & 44.31            & 44.76       & 44.86       & 17.48       & 42.18       \\ \hline
            \end{tabular}
    }
    \caption{Scores for Structured Answer Selection with modified MMLU-PRO Choices}
\end{table}

\begin{table}[]
    \centering
    \resizebox{\columnwidth}{!}{
        \begin{tabular}{c|c|c|c|c|c}
            \hline
            \textbf{models}                    & \textbf{Default} & \textbf{$\mathbf{v_{none}}$} & \textbf{$\mathbf{v_{all}}$} & \textbf{$\mathbf{v_{none-correct}}$} & \textbf{$\mathbf{v_{both}}$} \\ \hline
            google/gemma-2-2b                  & 49.59            & 52.92       & 22.81       & 13.19       & 16.09       \\
            google/gemma-2-2b-it               & 56.87            & 56.20       & 44.59       & 28.49       & 31.46       \\
            google/gemma-7b                    & 61.75            & 65.62       & 55.66       & 9.73        & 51.57       \\
            google/gemma-7b-it                 & 50.22            & 56.69       & 54.60       & 5.70        & 36.11       \\ \hline
            meta-llama/Llama-3.2-1B-Instruct   & 46.04            & 48.70       & 5.85        & 10.88       & 14.84       \\
            meta-llama/Llama-3.1-8B            & 63.44            & 68.37       & 61.45       & 5.40        & 49.96       \\
            meta-llama/Llama-3.1-8B-Instruct   & 67.95            & 71.10       & 62.72       & 14.84       & 54.02       \\
            meta-llama/Llama-2-13b-chat-hf     & 53.10            & 57.06       & 33.11       & 13.89       & 17.92       \\ \hline
            mistralai/Mistral-7B-v0.3          & 59.14            & 60.37       & 26.21       & 19.17       & 25.98       \\
            mistralai/Mistral-7B-Instruct-v0.3 & 59.63            & 62.32       & 41.59       & 20.27       & 30.05       \\ \hline
            microsoft/phi-4                    & 76.91            & 78.24       & 74.70       & 33.78       & 71.28       \\
            microsoft/Phi-3.5-mini-instruct    & 68.65            & 68.15       & 66.62       & 31.90       & 59.14       \\ \hline
            \end{tabular}
    }
    \caption{Scores for Structured Answer Selection with modified MMLU Choices}
\end{table}

\subsection{Plots showing percentage differnce across different benchmarks for Modified Questions}
\begin{figure}[h]
    \centering
    \includegraphics[width=\columnwidth]{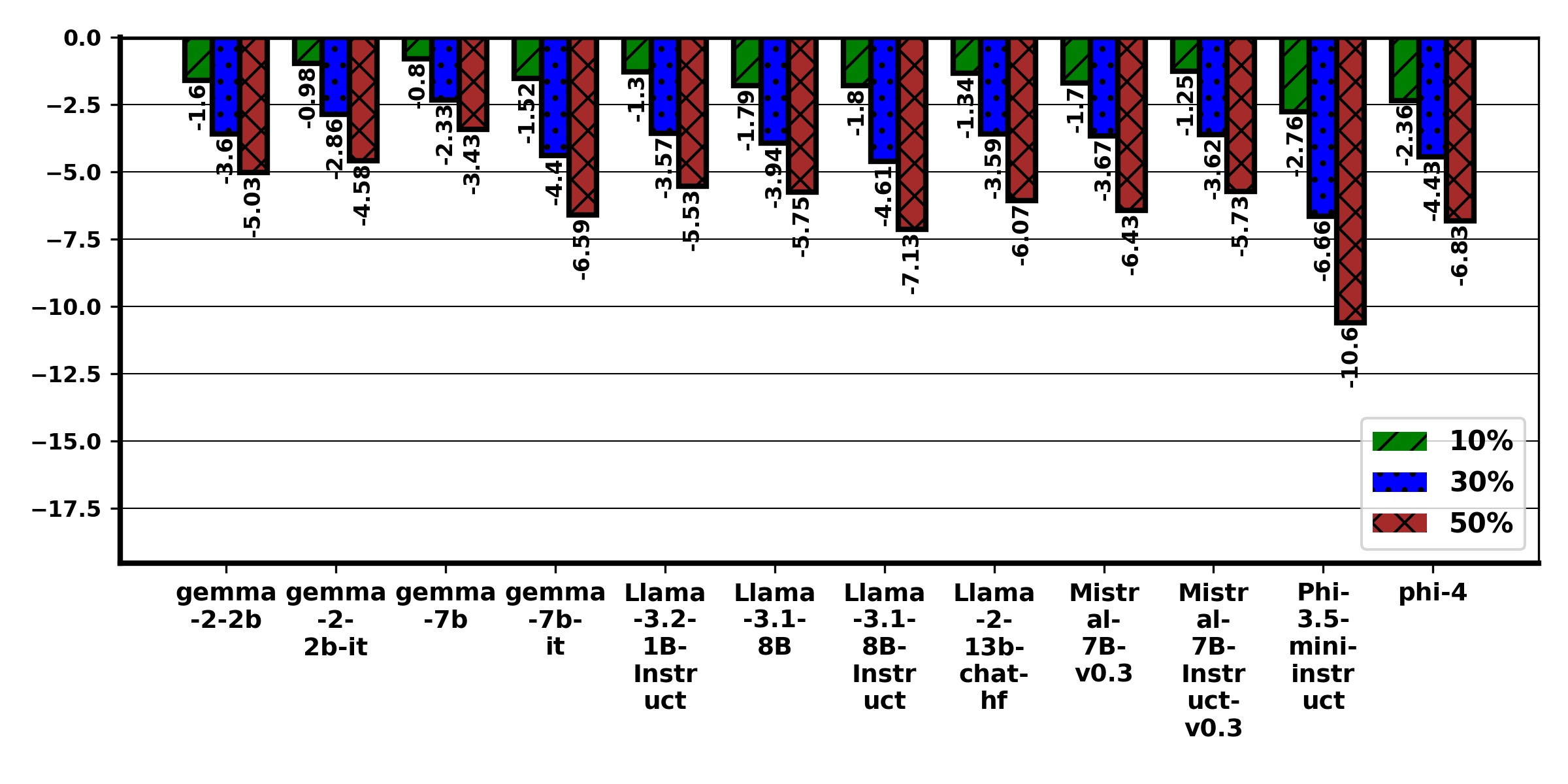}
    \caption{$\mathbf{v_0}$ results for MMLU benchmark across models.}
    \label{fig:results_1}
\end{figure}

\begin{figure}[h]
    \centering
    \includegraphics[width=\columnwidth]{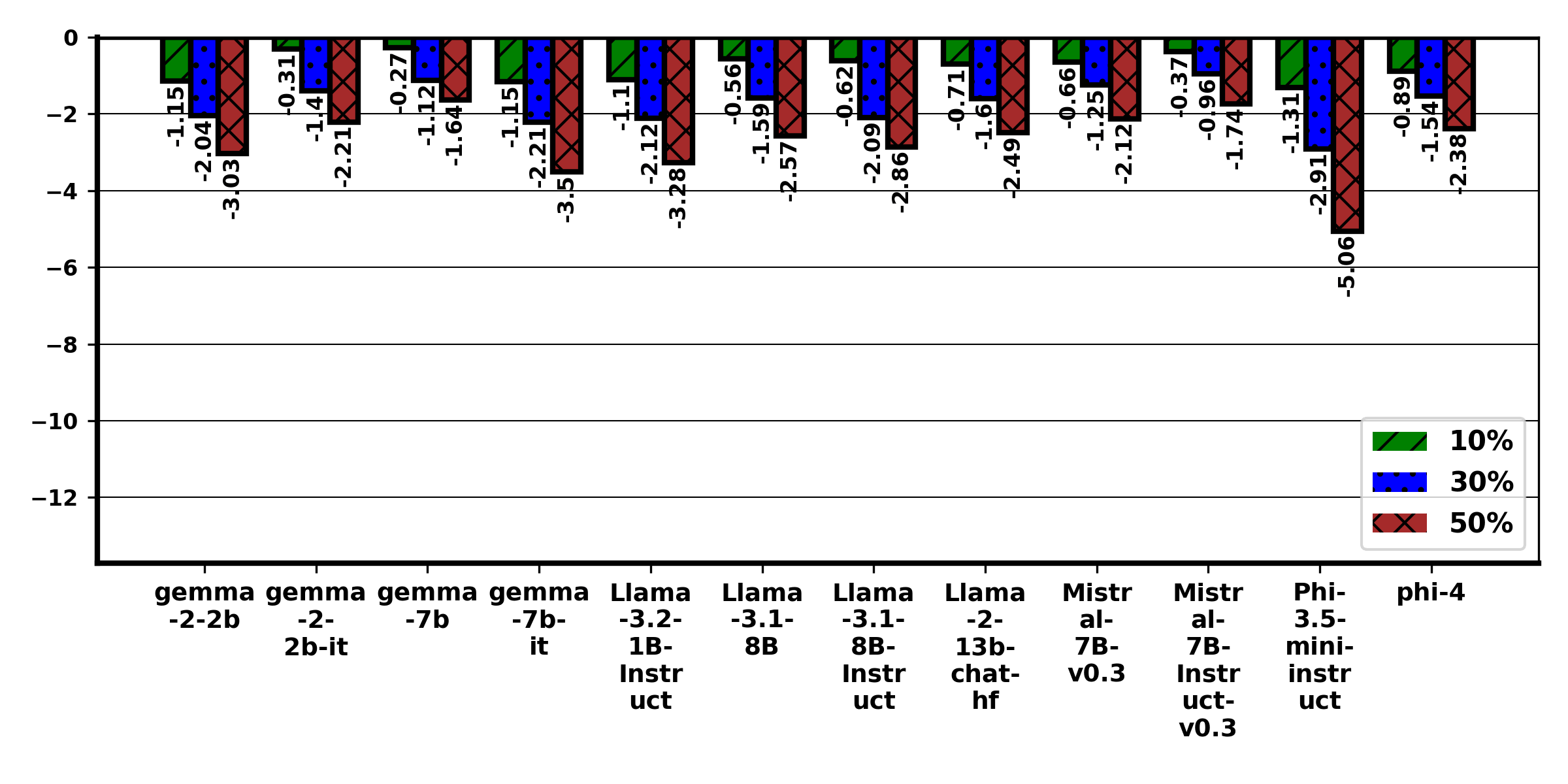}
    \caption{$\mathbf{v_1}$ results for MMLU benchmark across models.}
    \label{fig:results_1}
\end{figure}

\begin{figure}[h]
    \centering
    \includegraphics[width=\columnwidth]{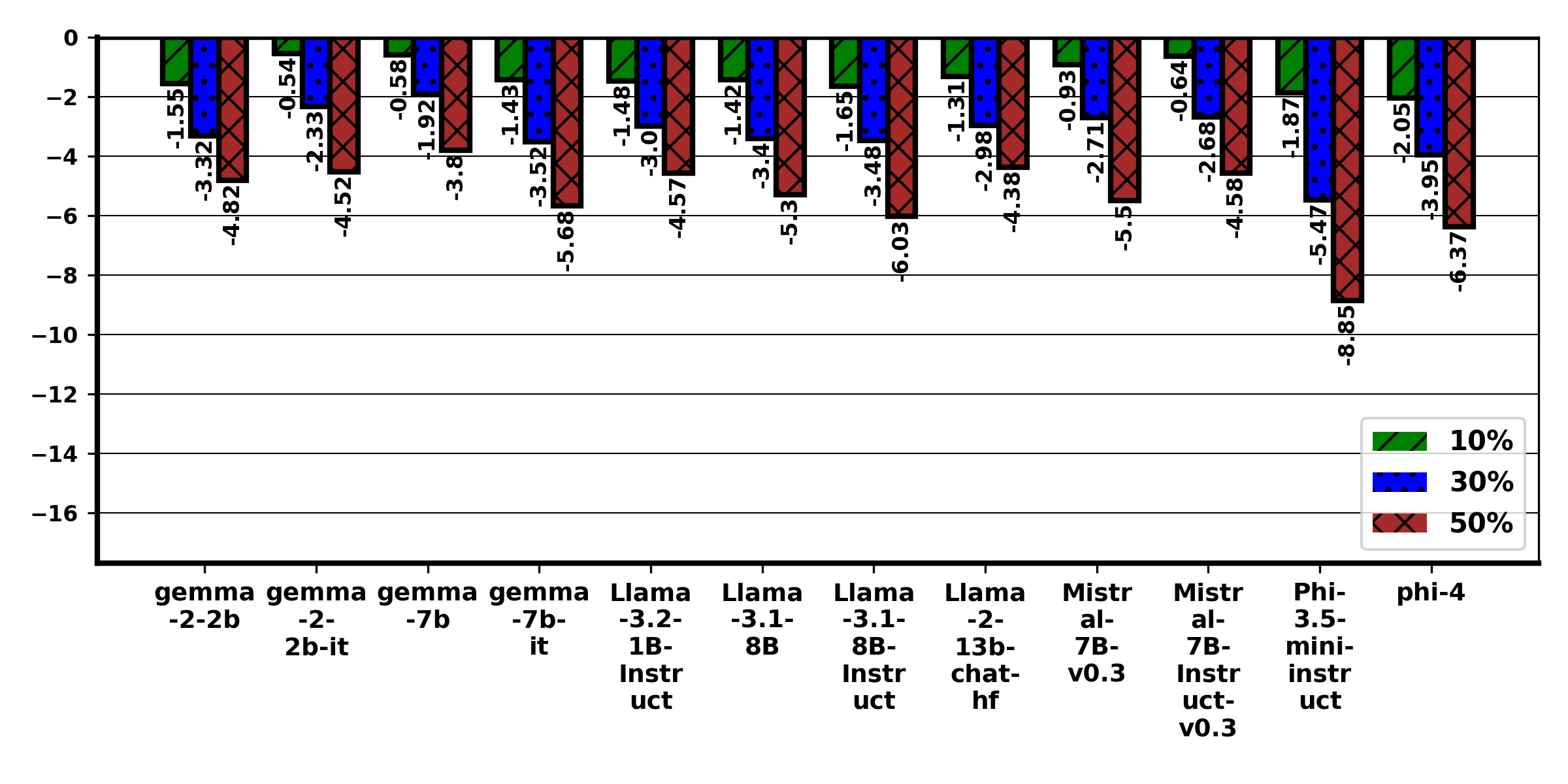}
    \caption{$\mathbf{v_2}$ results for MMLU benchmark across models.}
    \label{fig:results_1}
\end{figure}

\begin{figure}[h]
    \centering
    \includegraphics[width=\columnwidth]{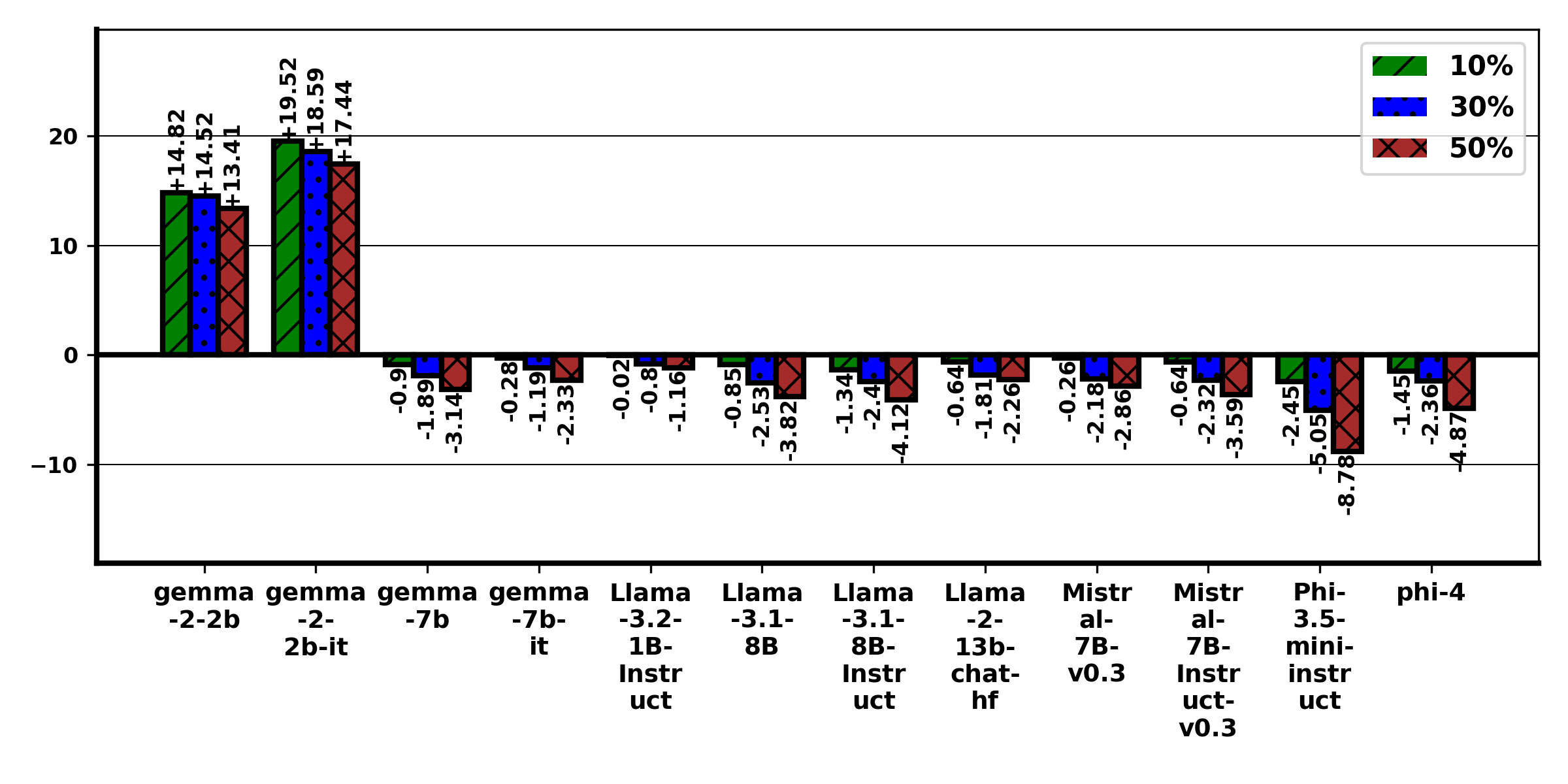}
    \caption{$\mathbf{v_0}$ results for MMLU-PRO benchmark across models.}
    \label{fig:results_1}
\end{figure}

\begin{figure}[h]
    \centering
    \includegraphics[width=\columnwidth]{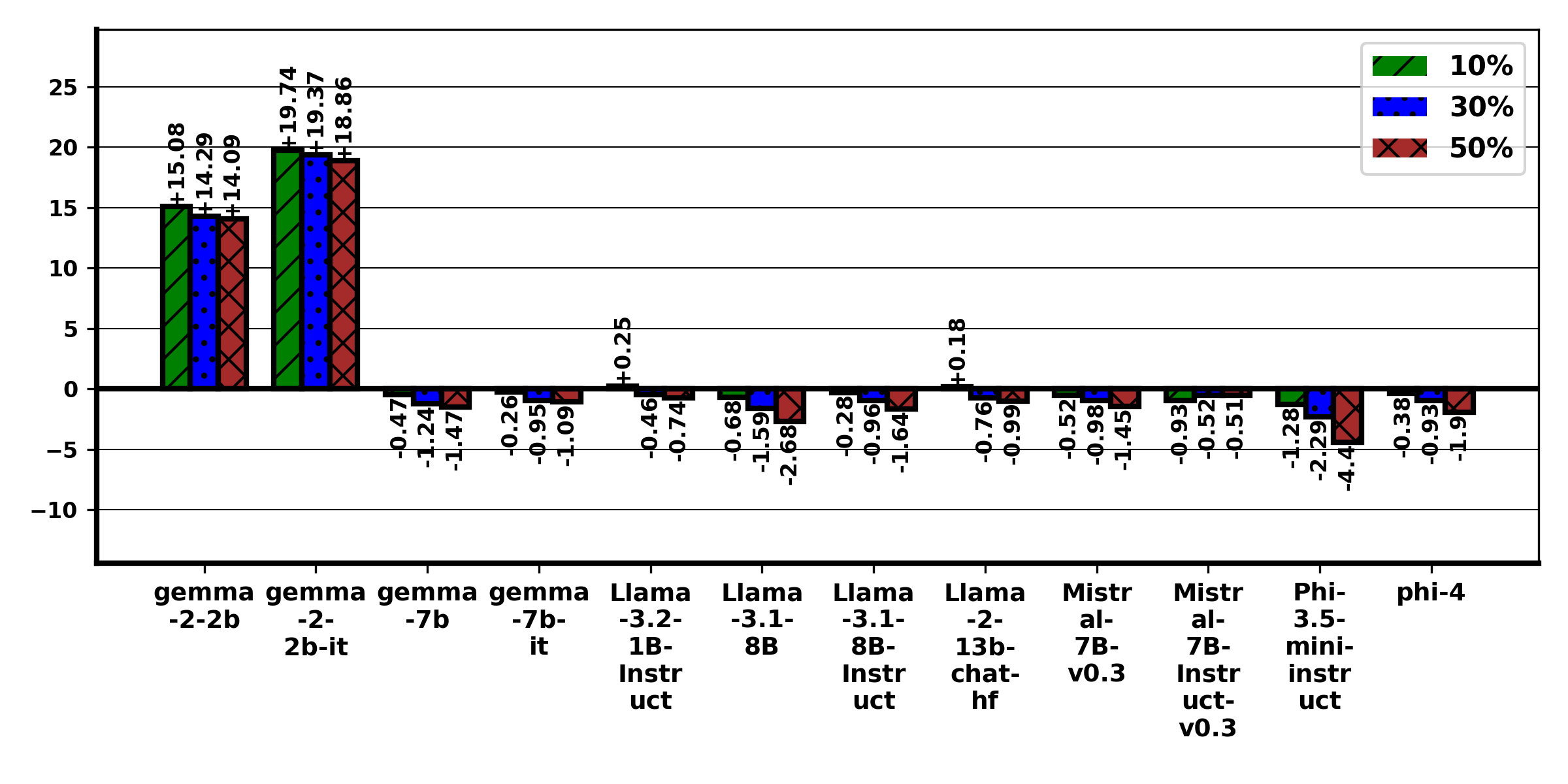}
    \caption{$\mathbf{v_1}$ results for MMLU-PRO benchmark across models.}
    \label{fig:results_1}
\end{figure}

\begin{figure}[h]
    \centering
    \includegraphics[width=\columnwidth]{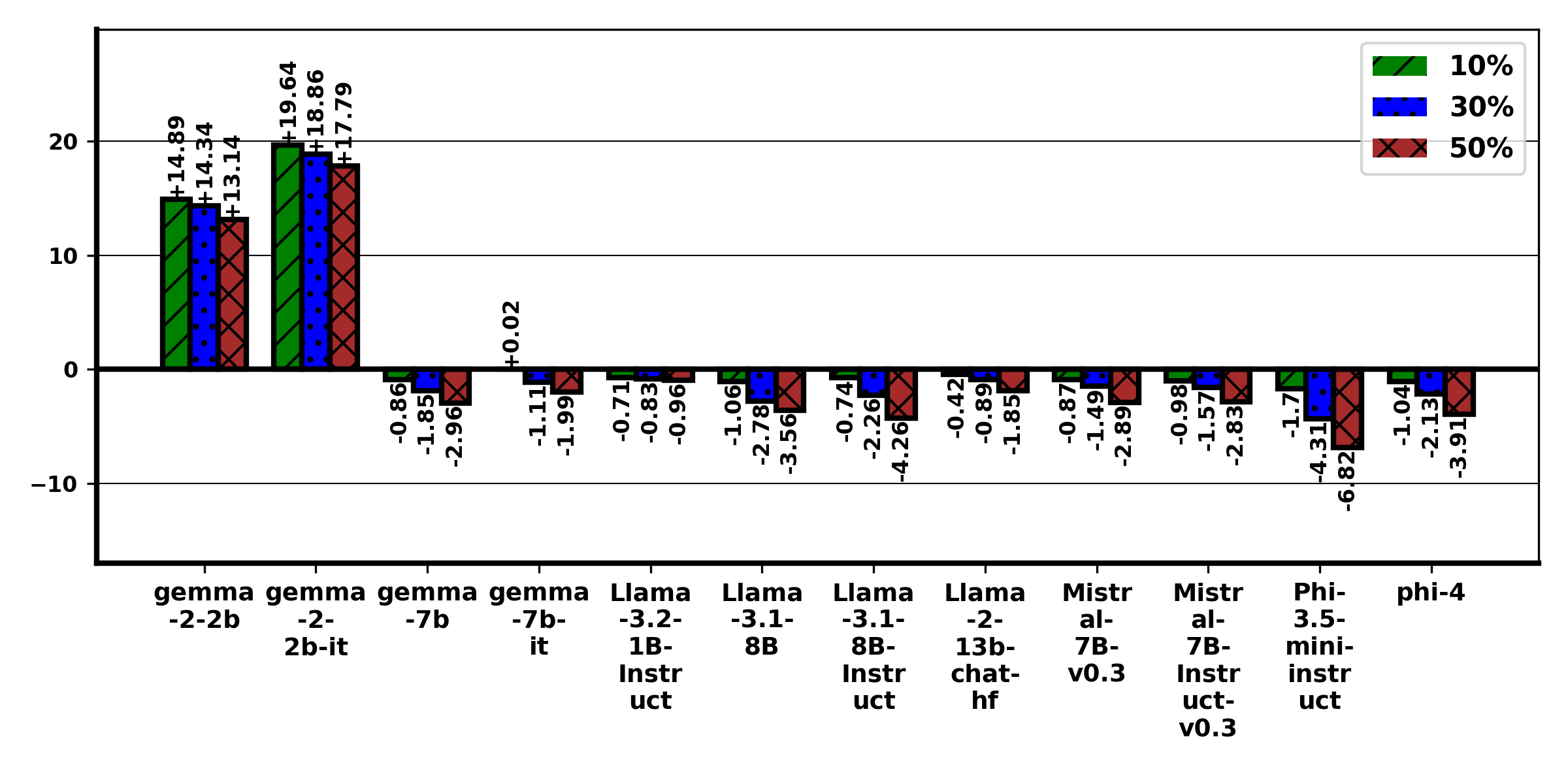}
    \caption{$\mathbf{v_2}$ results for MMLU-PRO benchmark across models.}
    \label{fig:results_1}
\end{figure}

\begin{figure}[h]
    \centering
    \includegraphics[width=\columnwidth]{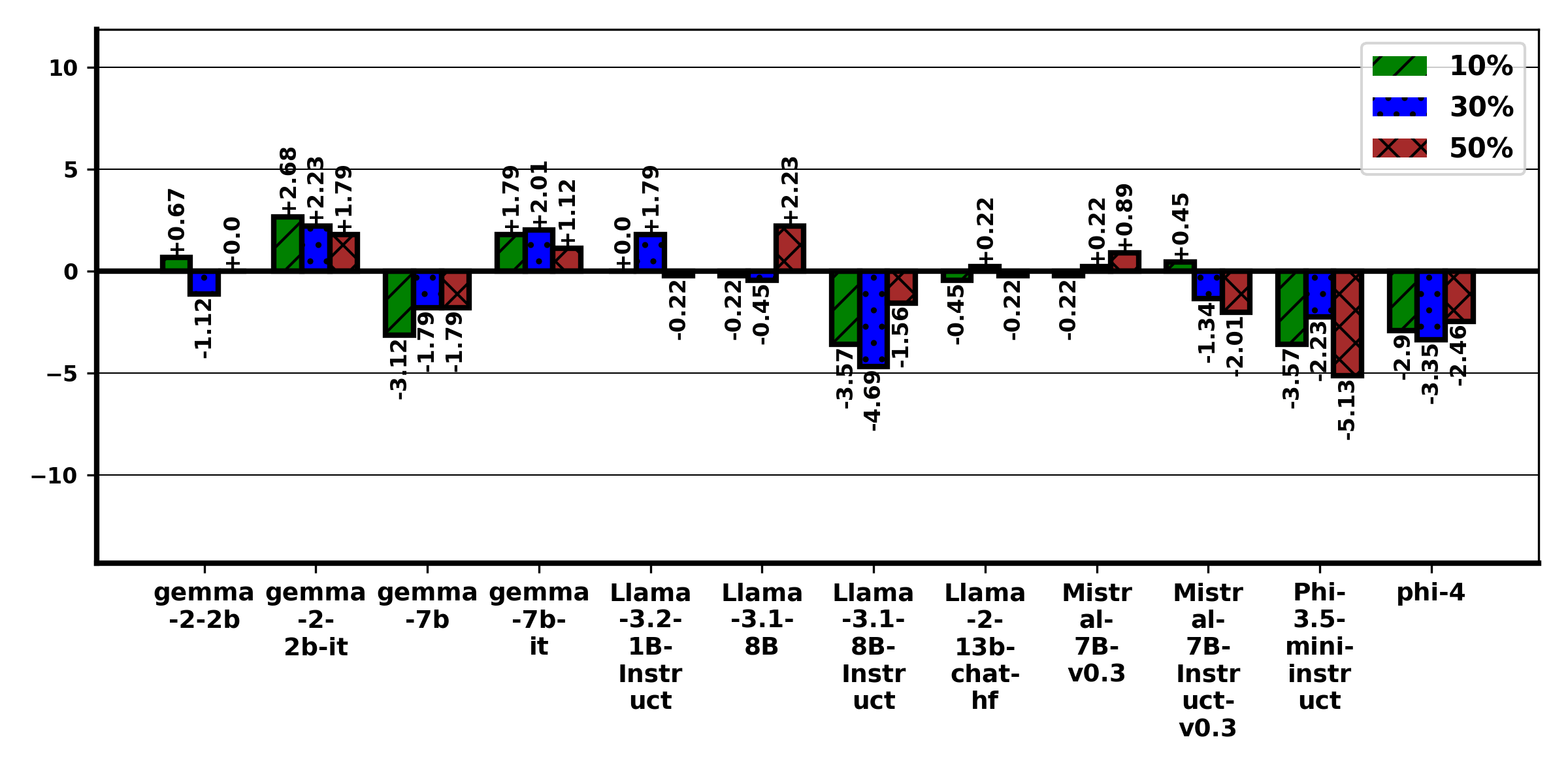}
    \caption{$\mathbf{v_0}$ results for GPQA benchmark across models.}
    \label{fig:results_1}
\end{figure}

\begin{figure}[h]
    \centering
    \includegraphics[width=\columnwidth]{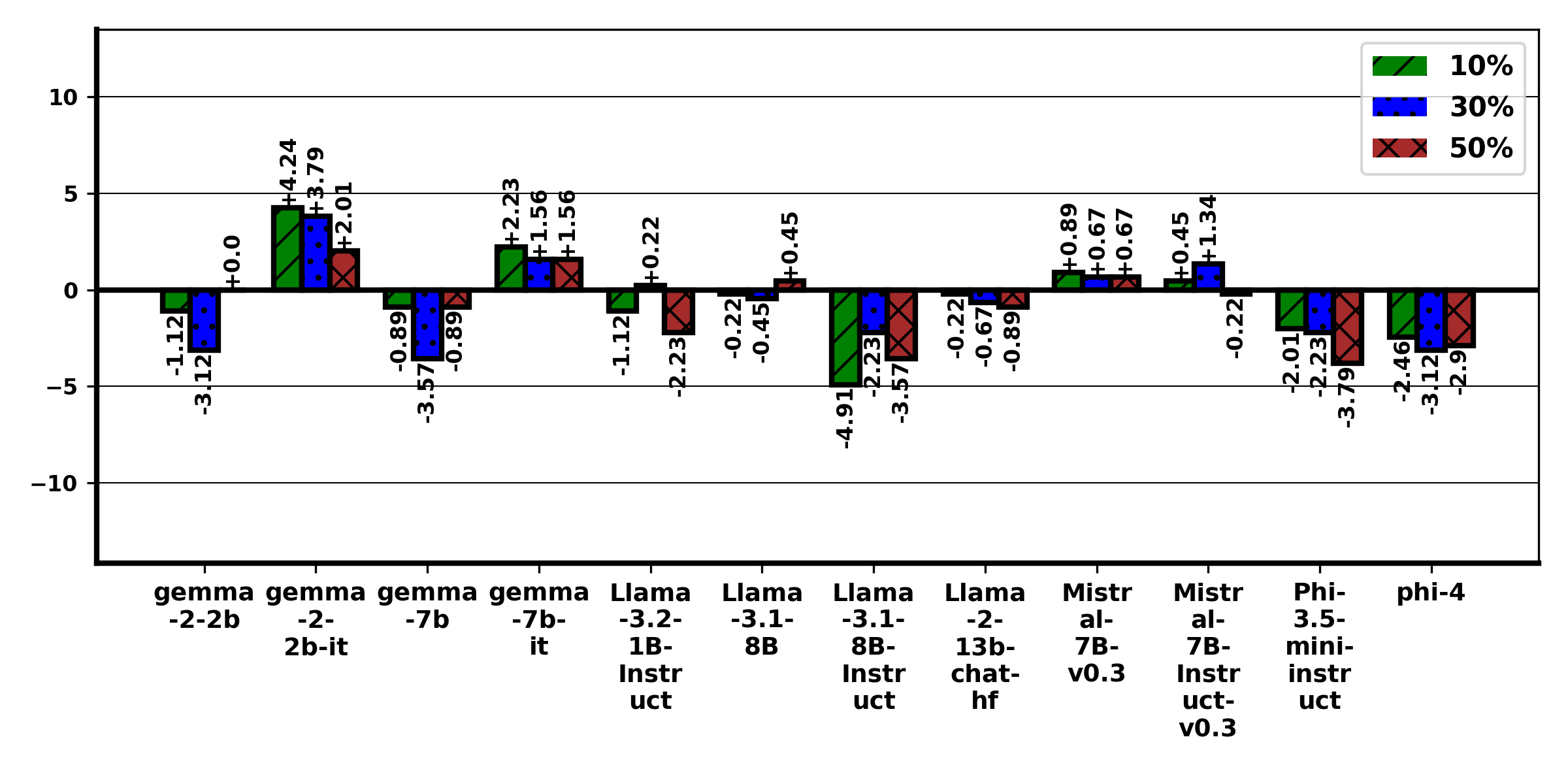}
    \caption{$\mathbf{v_1}$ results for GPQA benchmark across models.}
    \label{fig:results_1}
\end{figure}

\begin{figure}[h]
    \centering
    \includegraphics[width=\columnwidth]{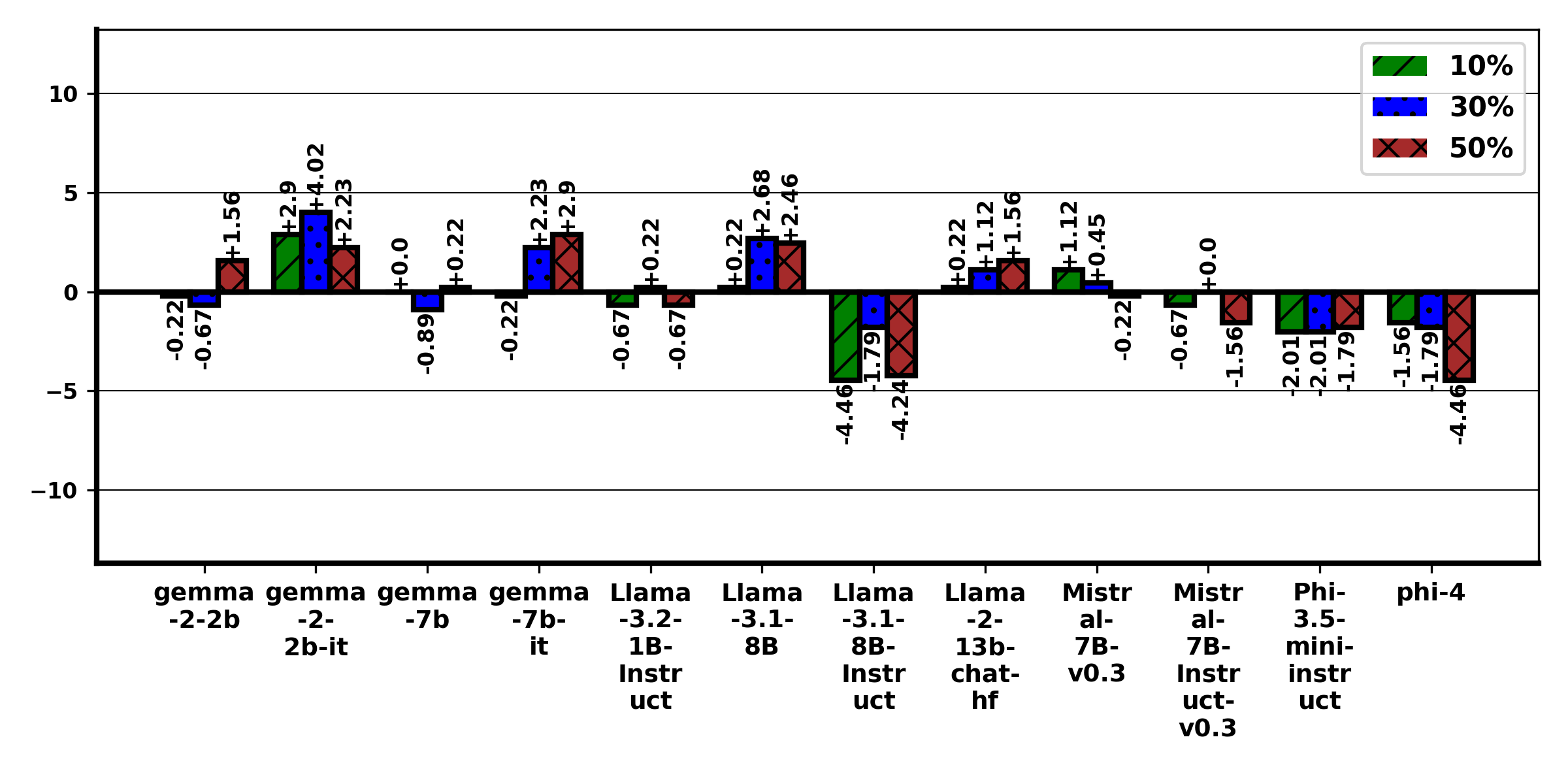}
    \caption{$\mathbf{v_2}$ results for GPQA benchmark across models.}
    \label{fig:results_1}
\end{figure}

\begin{figure}[h]
    \centering
    \includegraphics[width=\columnwidth]{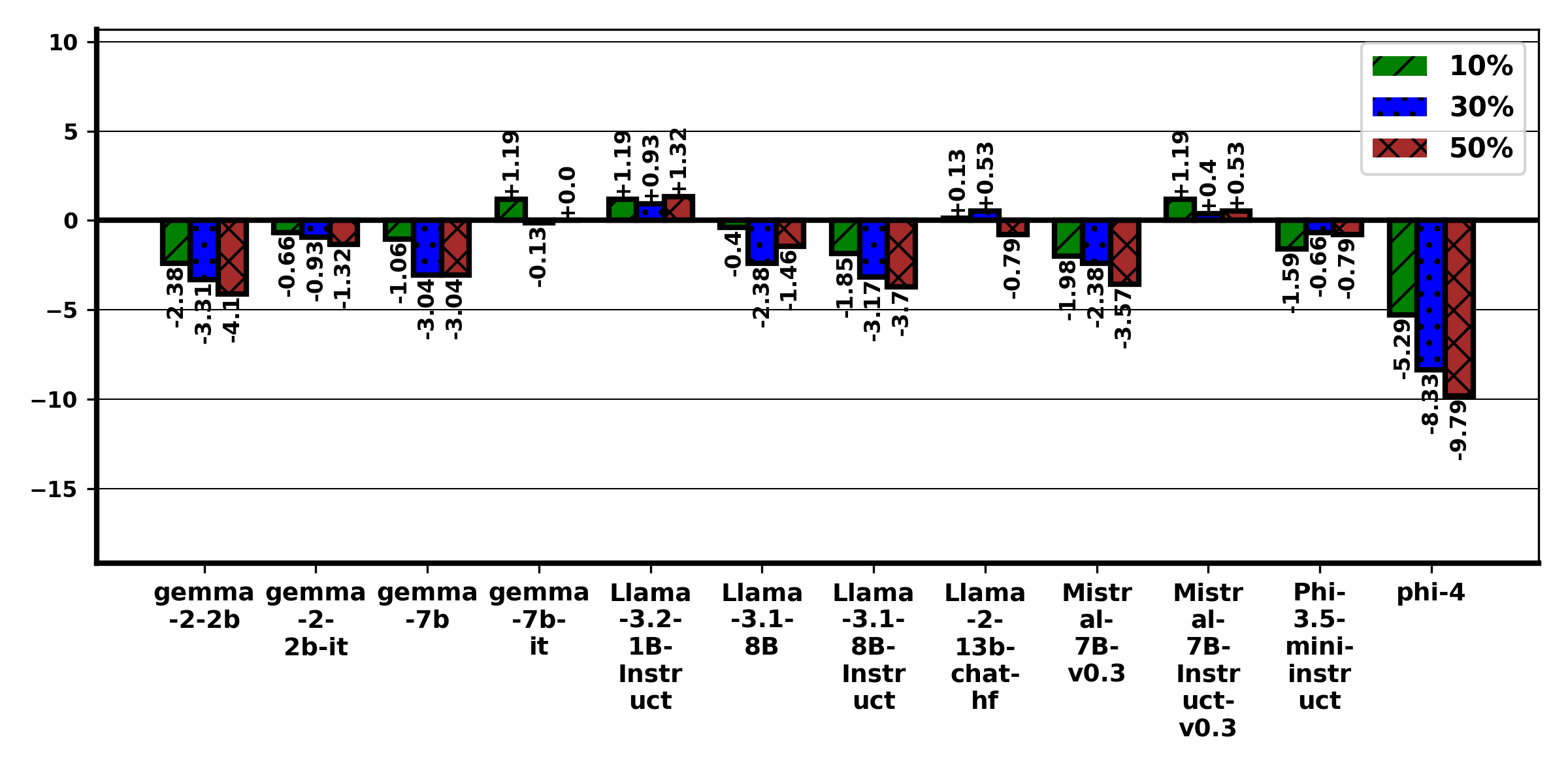}
    \caption{$\mathbf{v_0}$ results for MUSR benchmark across models.}
    \label{fig:results_1}
\end{figure}

\begin{figure}[h]
    \centering
    \includegraphics[width=\columnwidth]{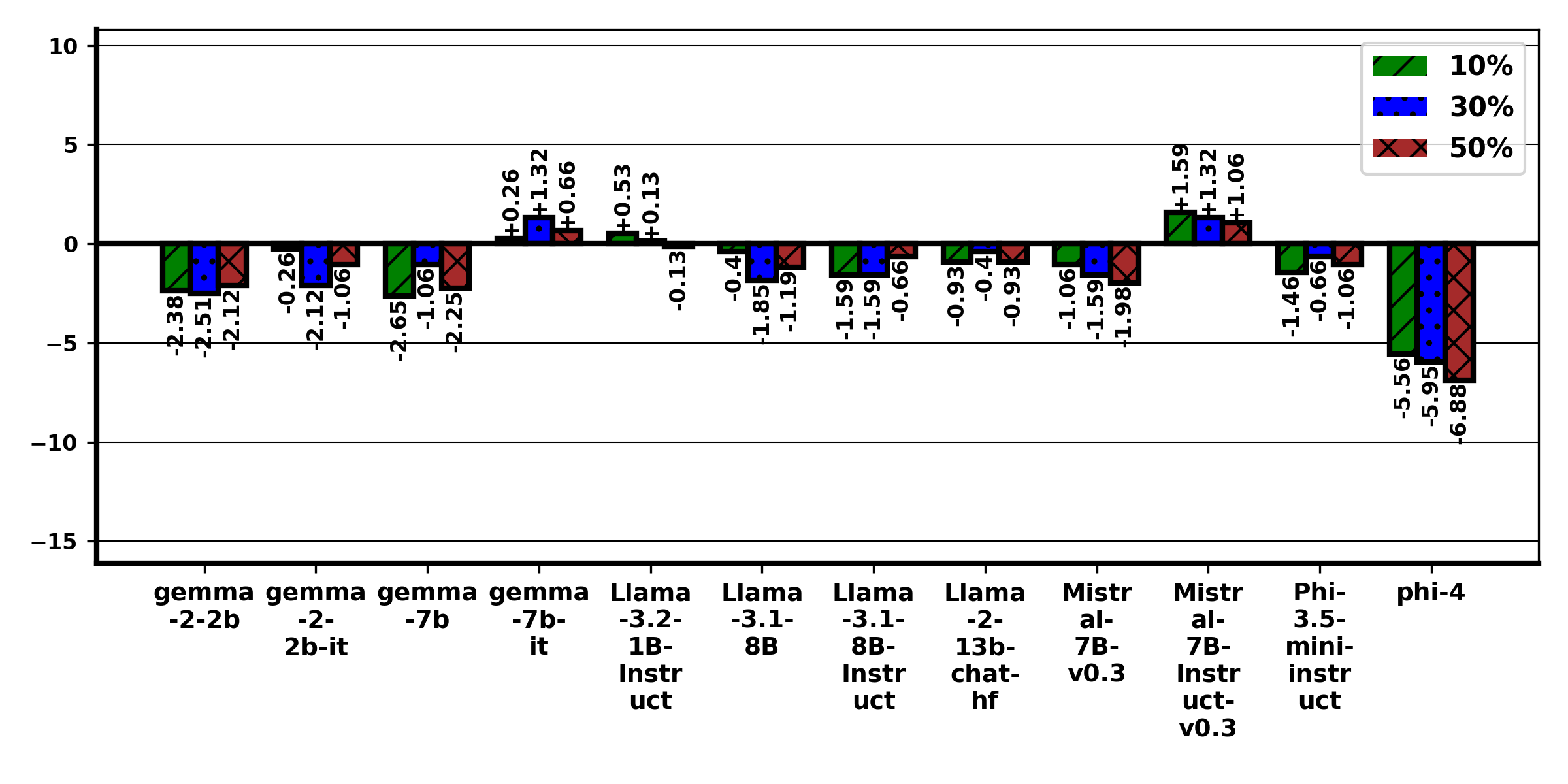}
    \caption{$\mathbf{v_1}$ results for MUSR benchmark across models.}
    \label{fig:results_1}
\end{figure}

\begin{figure}[h]
    \centering
    \includegraphics[width=\columnwidth]{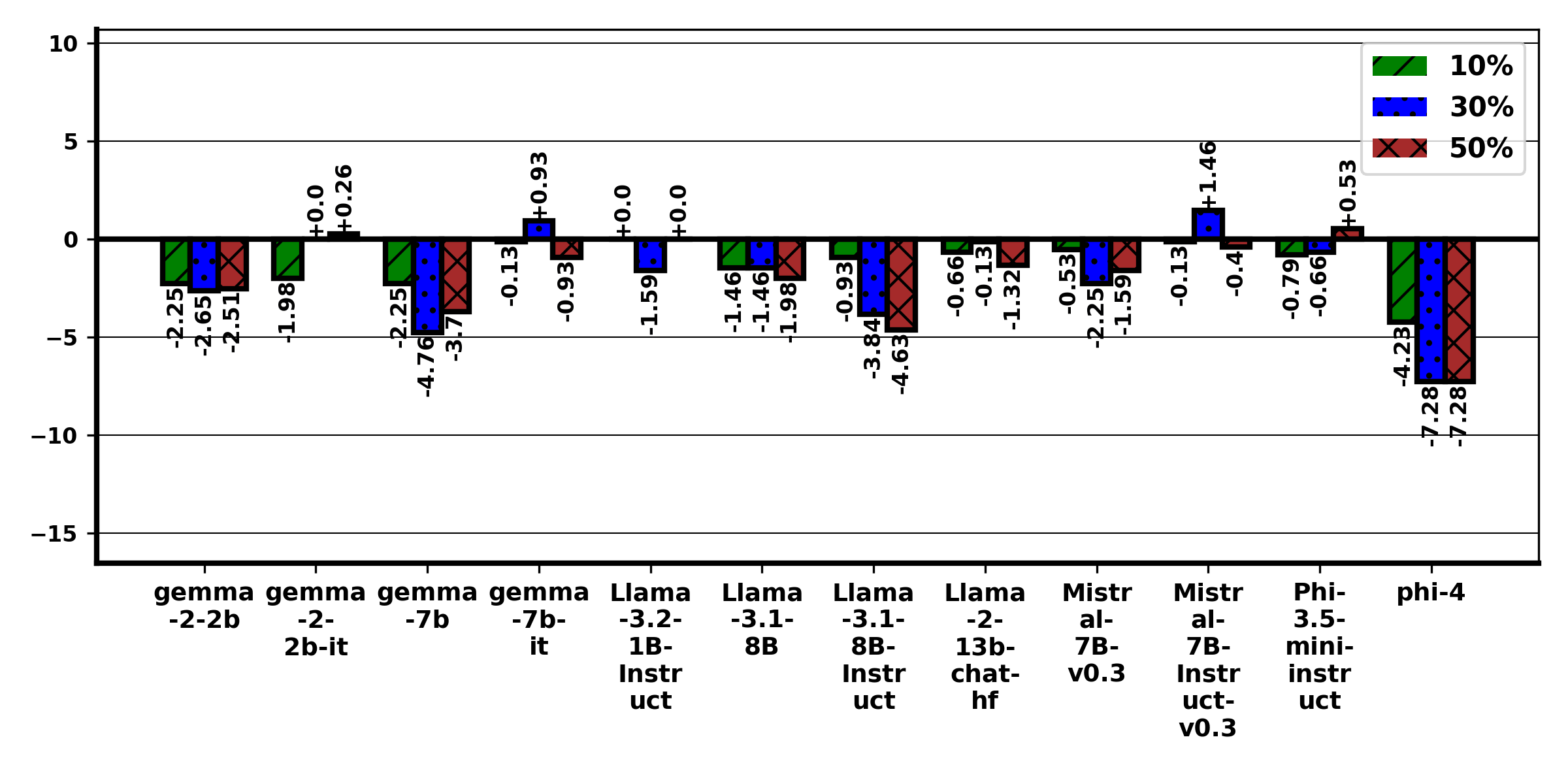}
    \caption{$\mathbf{v_2}$ results for MUSR benchmark across models.}
    \label{fig:results_1}
\end{figure}

\subsection{Plots showing comparision with default scores across different benchmarks for Modified Choices}
\begin{figure}[h]
    \centering
    \includegraphics[width=\columnwidth]{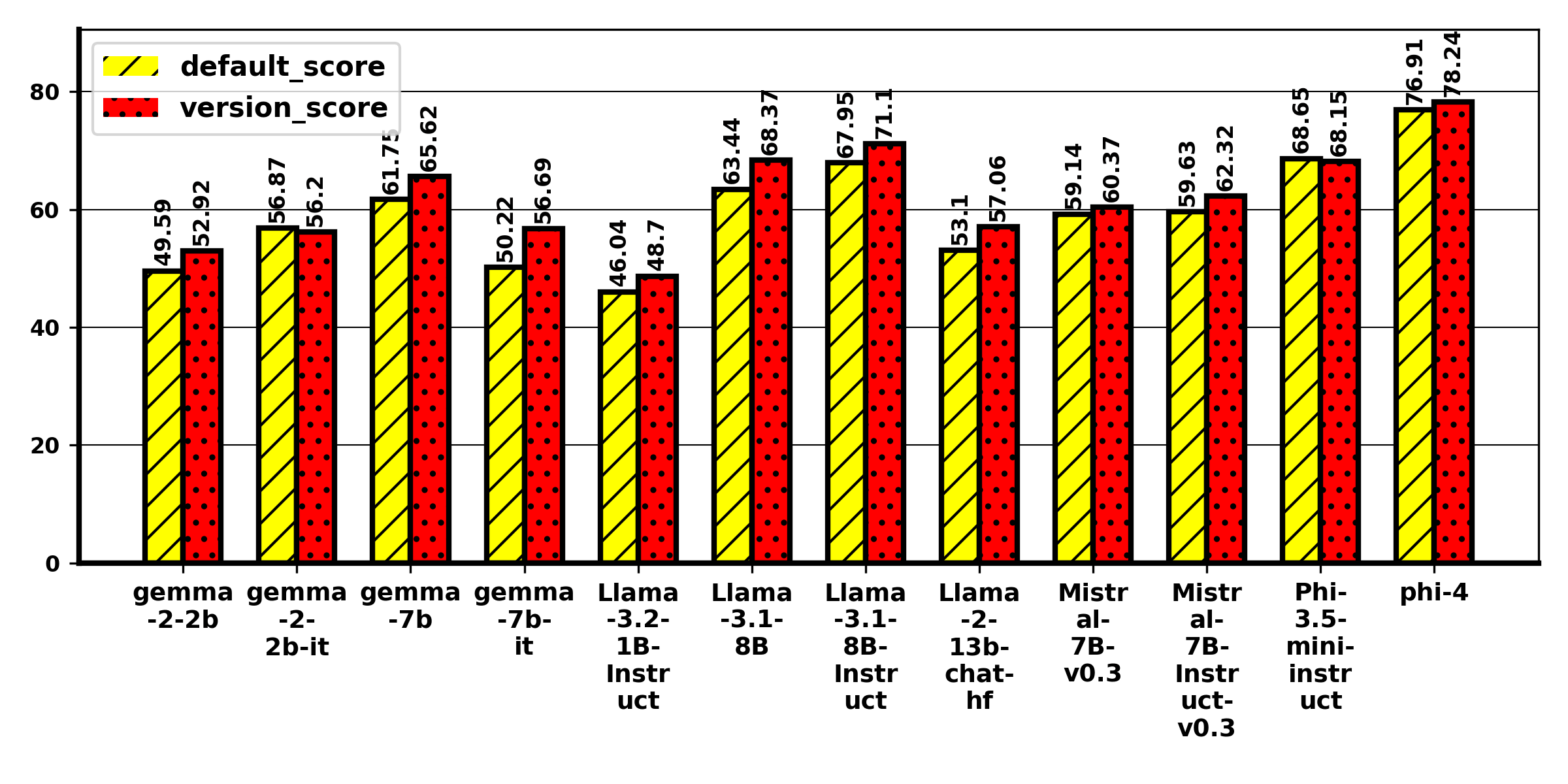}
    \caption{$\mathbf{v_{none}}$ results for MMLU benchmark across models.}
    \label{fig:results_1}
\end{figure}

\begin{figure}[h]
    \centering
    \includegraphics[width=\columnwidth]{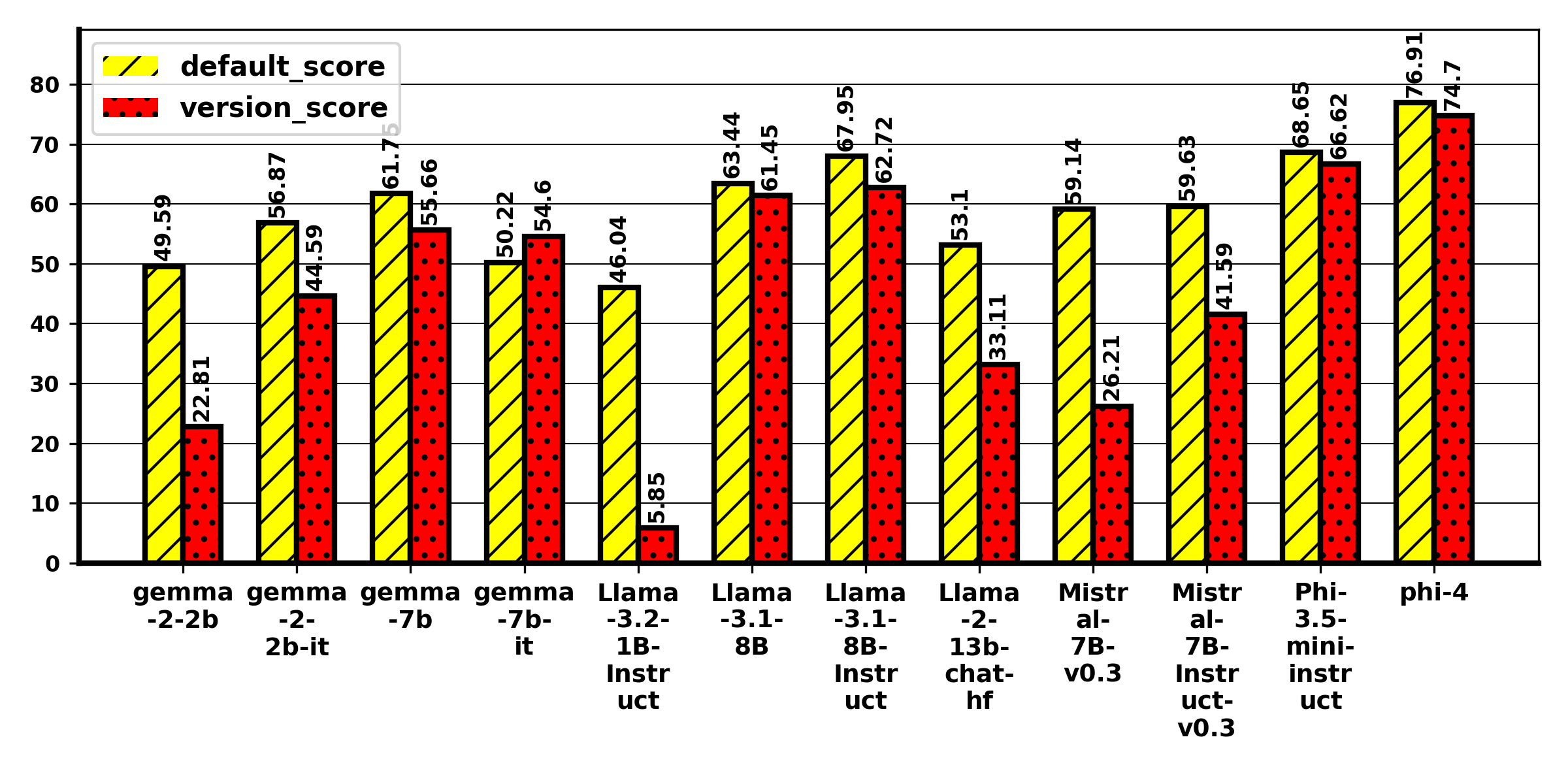}
    \caption{$\mathbf{v_{all}}$ results for MMLU benchmark across models.}
    \label{fig:results_1}
\end{figure} 

\begin{figure}[h]
    \centering
    \includegraphics[width=\columnwidth]{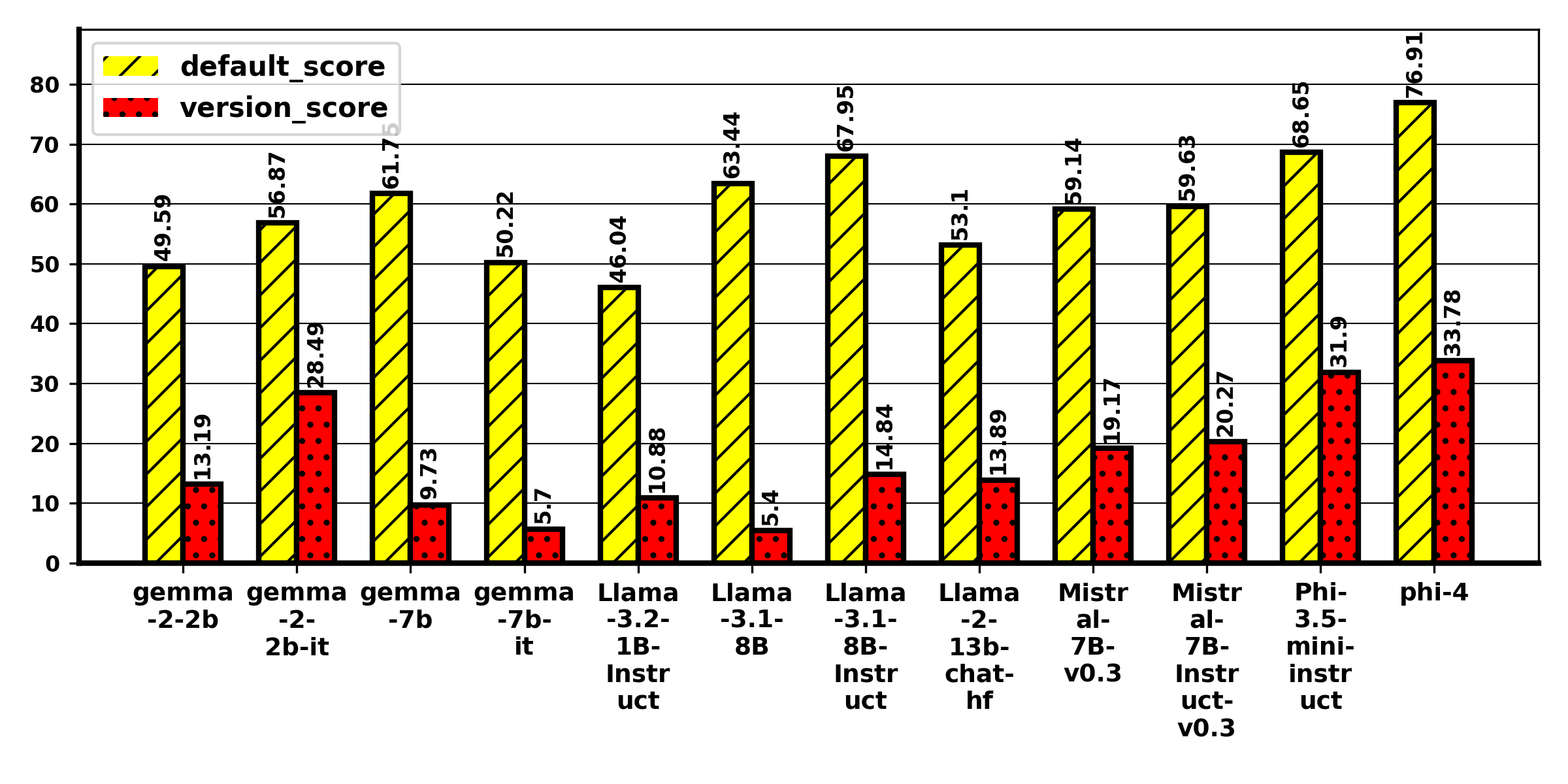}
    \caption{$\mathbf{v_{none-correct}}$ results for MMLU benchmark across models.}
    \label{fig:results_1}
\end{figure}

\begin{figure}[h]
    \centering
    \includegraphics[width=\columnwidth]{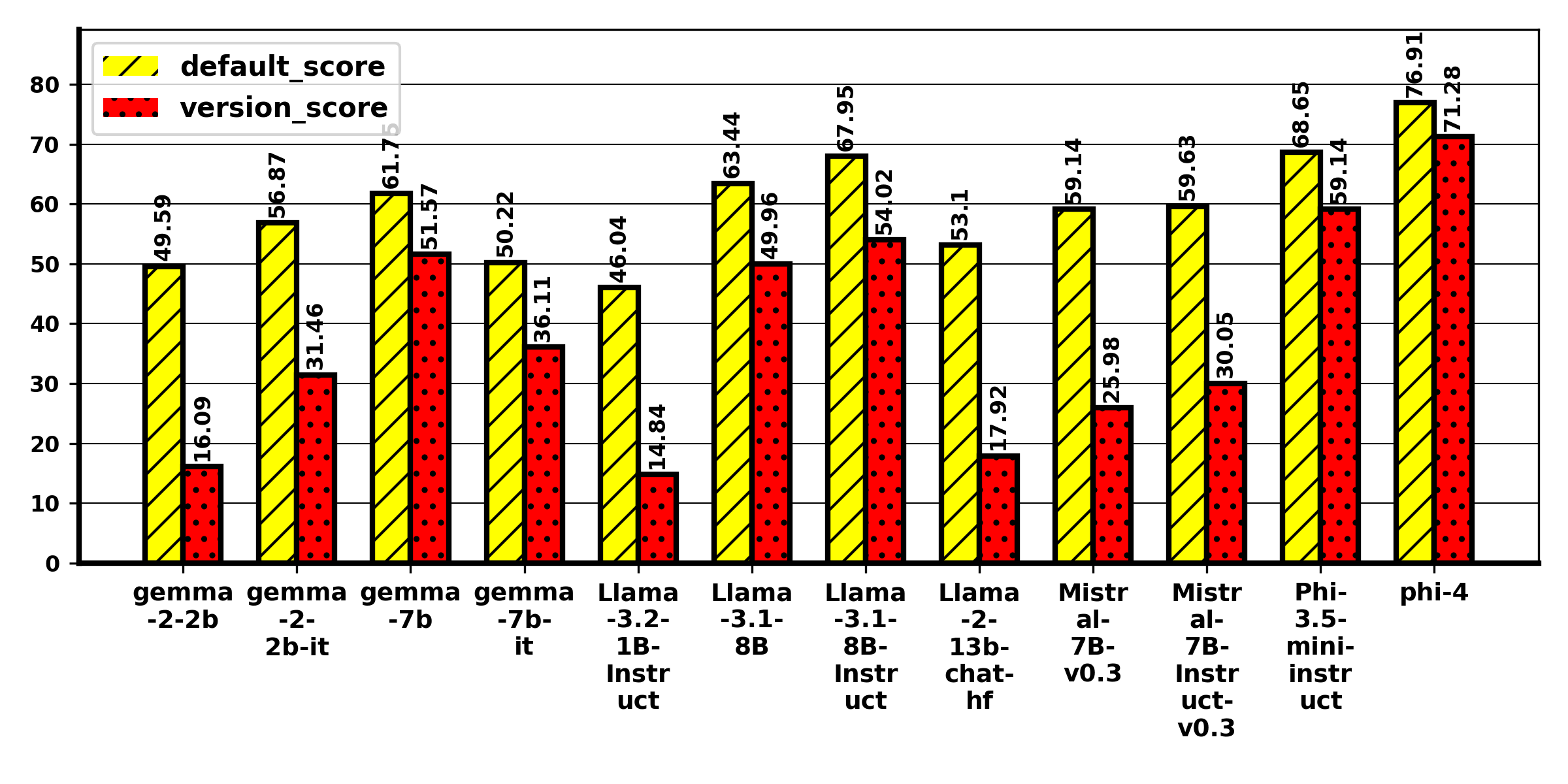}
    \caption{$\mathbf{v_{both}}$ results for MMLU benchmark across models.}
    \label{fig:results_1}
\end{figure}

\begin{figure}[h]
    \centering
    \includegraphics[width=\columnwidth]{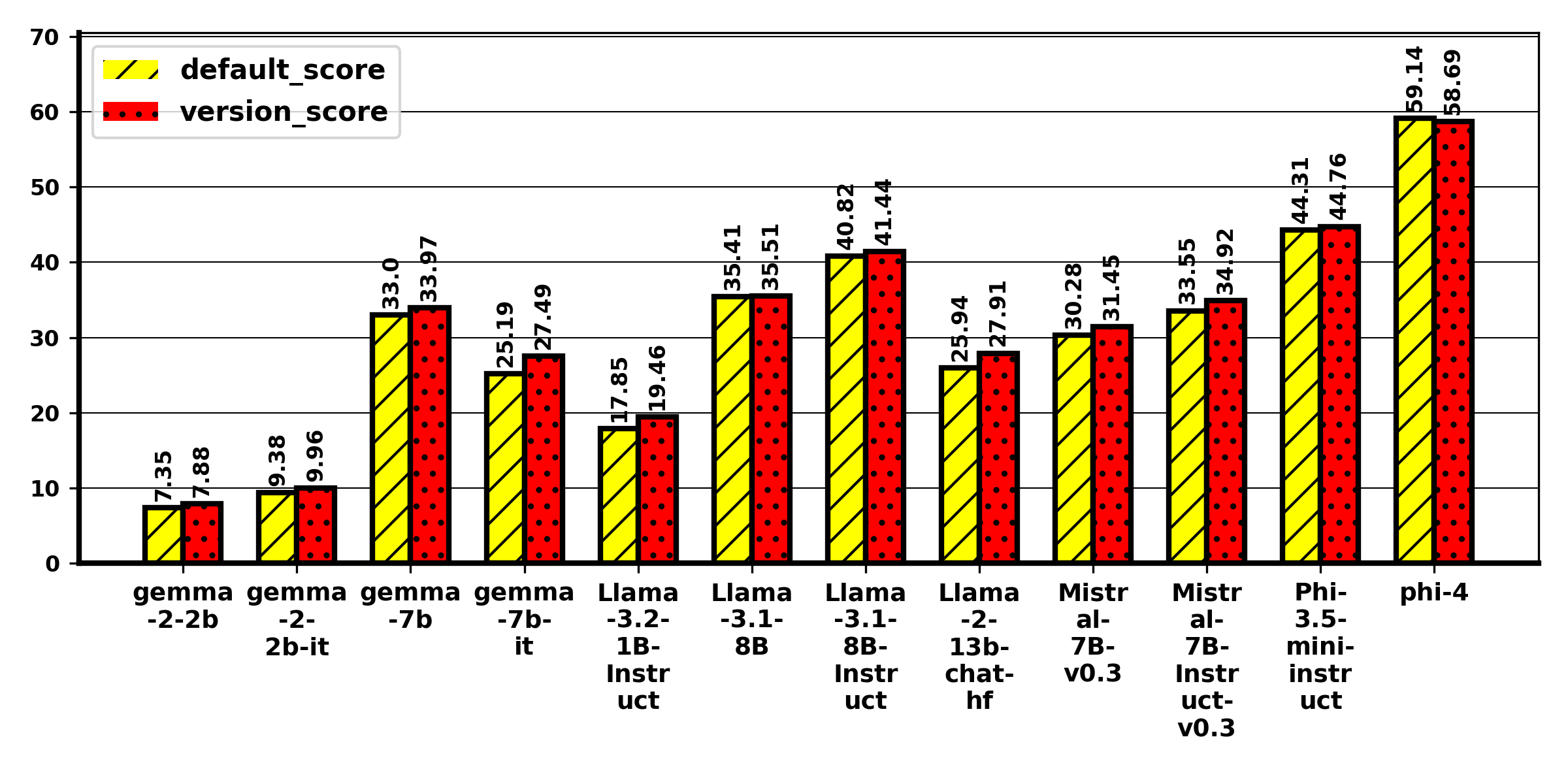}
    \caption{$\mathbf{v_{none}}$ results for MMLU-PRO benchmark across models.}
    \label{fig:results_1}
\end{figure}

\begin{figure}[h]
    \centering
    \includegraphics[width=\columnwidth]{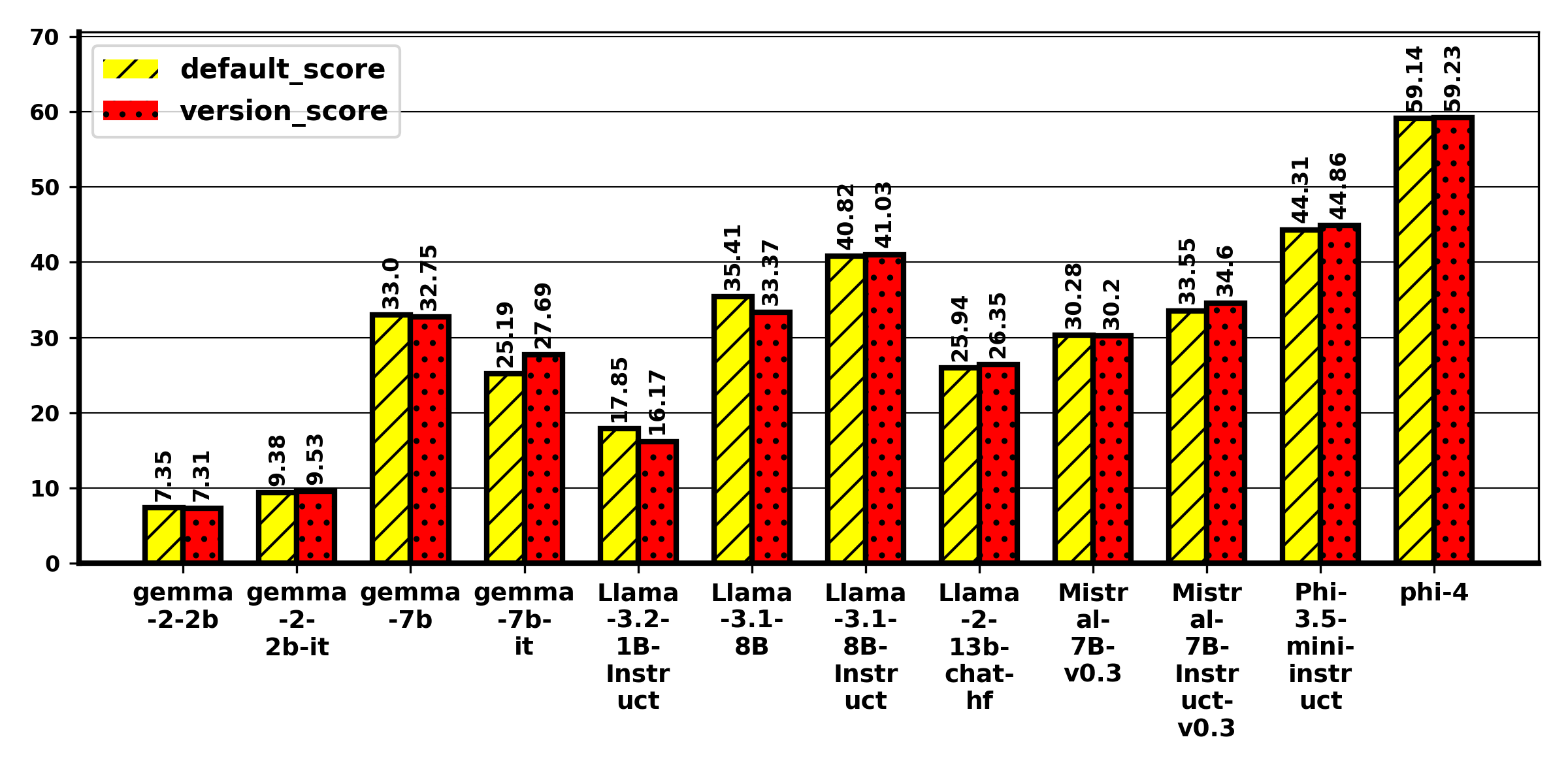}
    \caption{$\mathbf{v_{all}}$ results for MMLU-PRO benchmark across models.}
    \label{fig:results_1}
\end{figure}

\begin{figure}[h]
    \centering
    \includegraphics[width=\columnwidth]{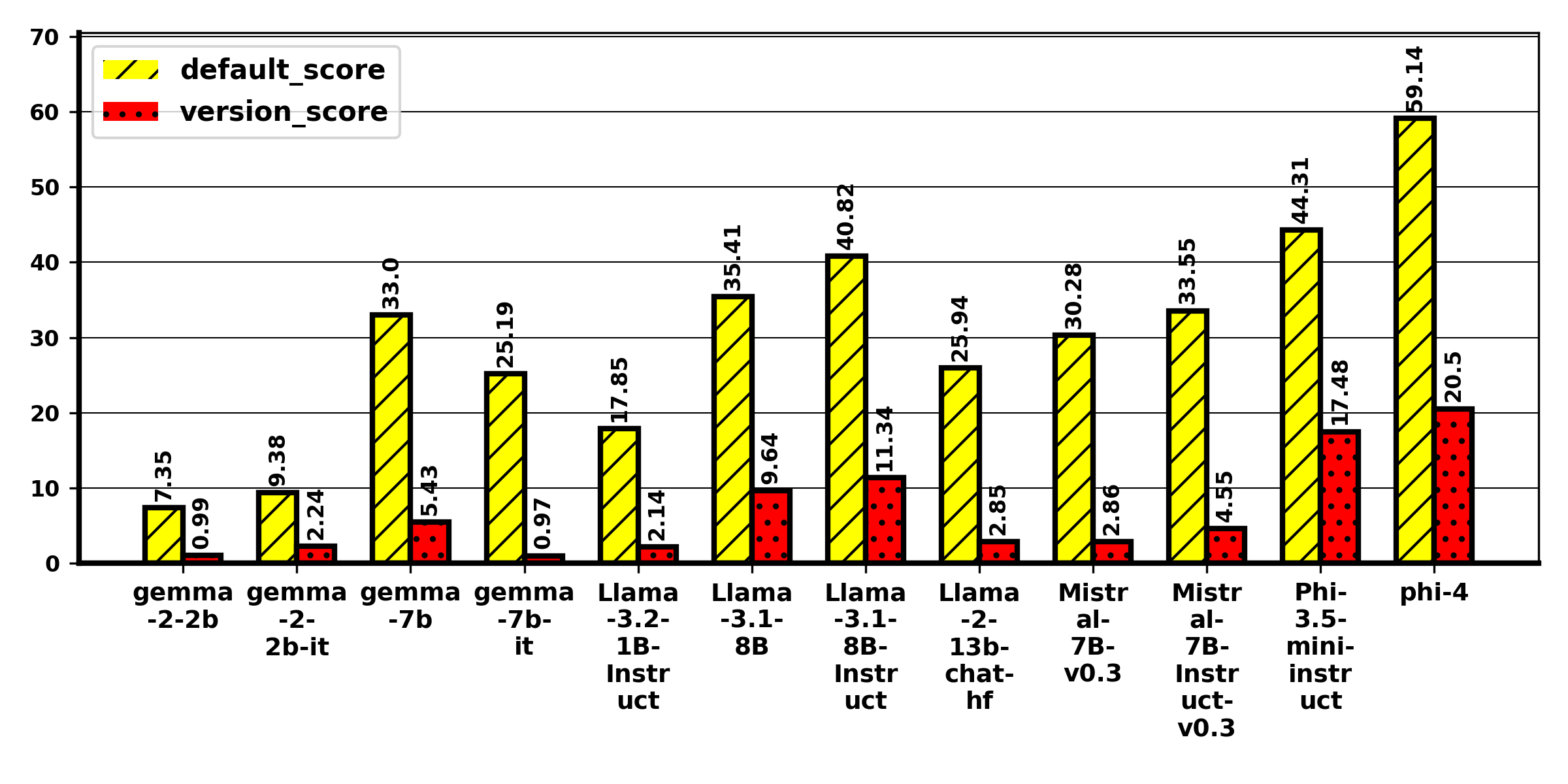}
    \caption{$\mathbf{v_{none-correct}}$ results for MMLU-PRO benchmark across models.}
    \label{fig:results_1}
\end{figure}

\begin{figure}[h]
    \centering
    \includegraphics[width=\columnwidth]{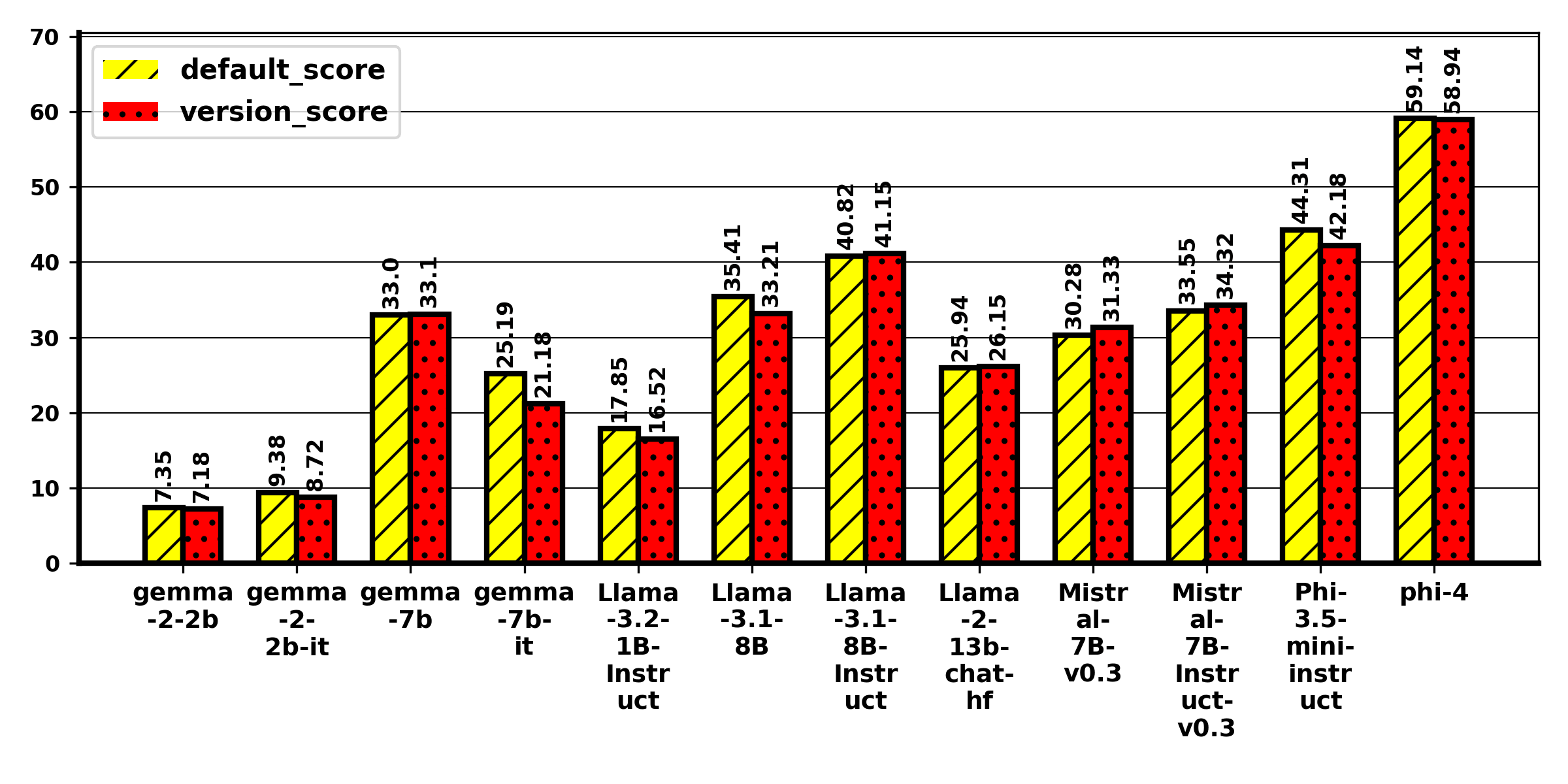}
    \caption{$\mathbf{v_{both}}$ results for MMLU-PRO benchmark across models.}
    \label{fig:results_1}
\end{figure}

\begin{figure}[h]
    \centering
    \includegraphics[width=\columnwidth]{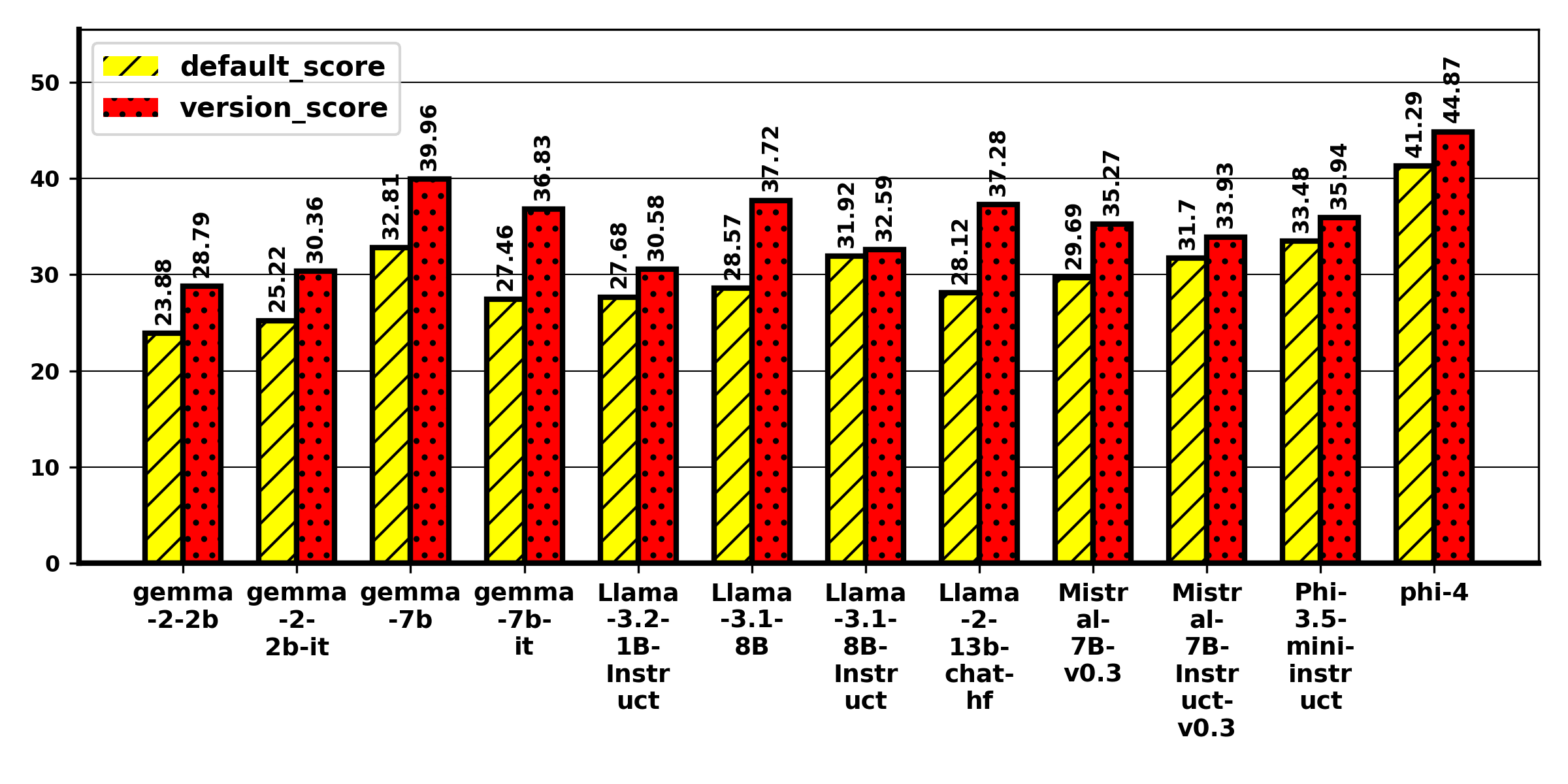}
    \caption{$\mathbf{v_{none}}$ results for GPQA benchmark across models.}
    \label{fig:results_1}
\end{figure}

\begin{figure}[h]
    \centering
    \includegraphics[width=\columnwidth]{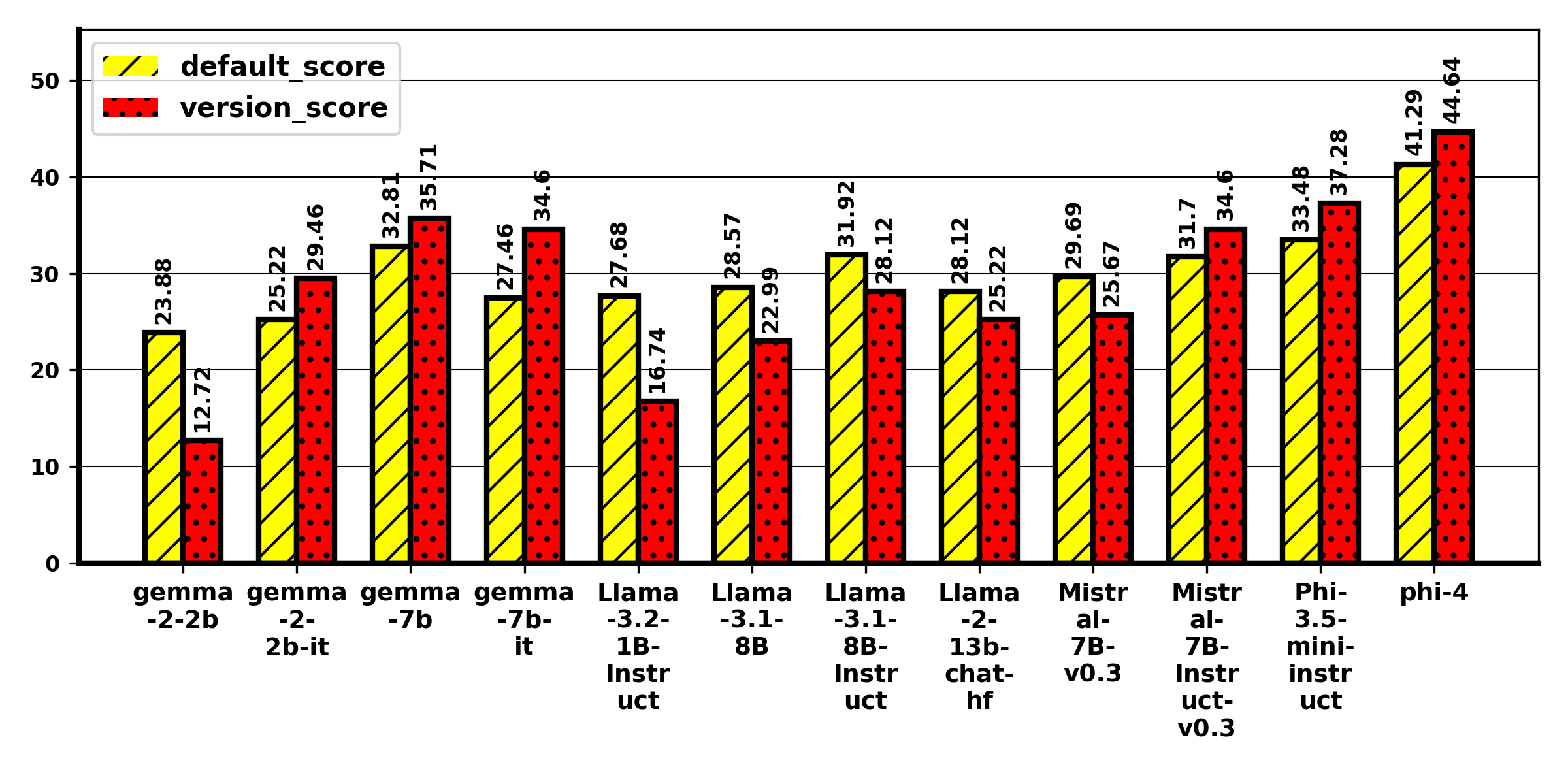}
    \caption{$\mathbf{v_{all}}$ results for GPQA benchmark across models.}
    \label{fig:results_1}
\end{figure}

\begin{figure}[h]
    \centering
    \includegraphics[width=\columnwidth]{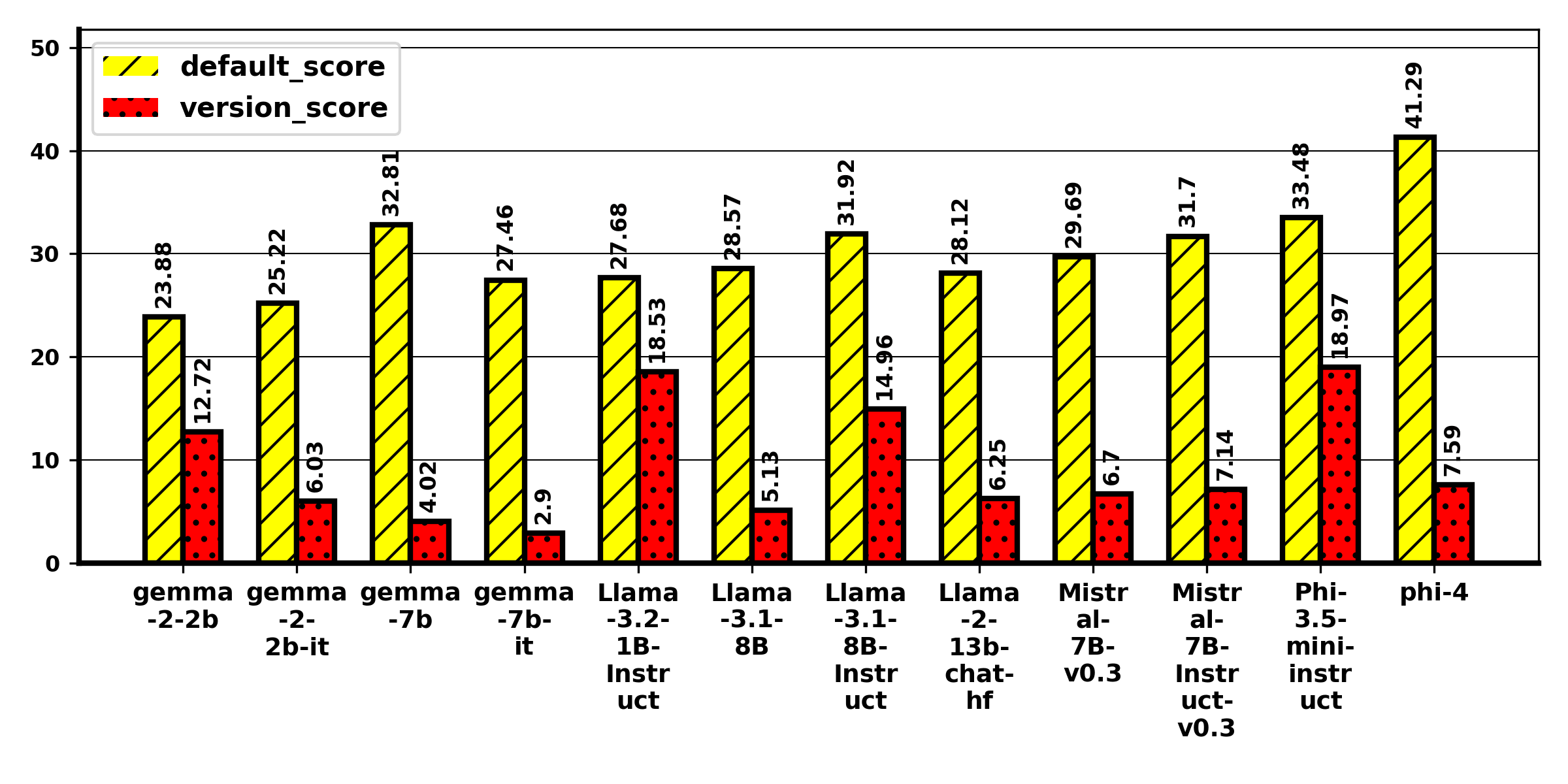}
    \caption{$\mathbf{v_{none-correct}}$ results for GPQA benchmark across models.}
    \label{fig:results_1}
\end{figure}

\begin{figure}[h]
    \centering
    \includegraphics[width=\columnwidth]{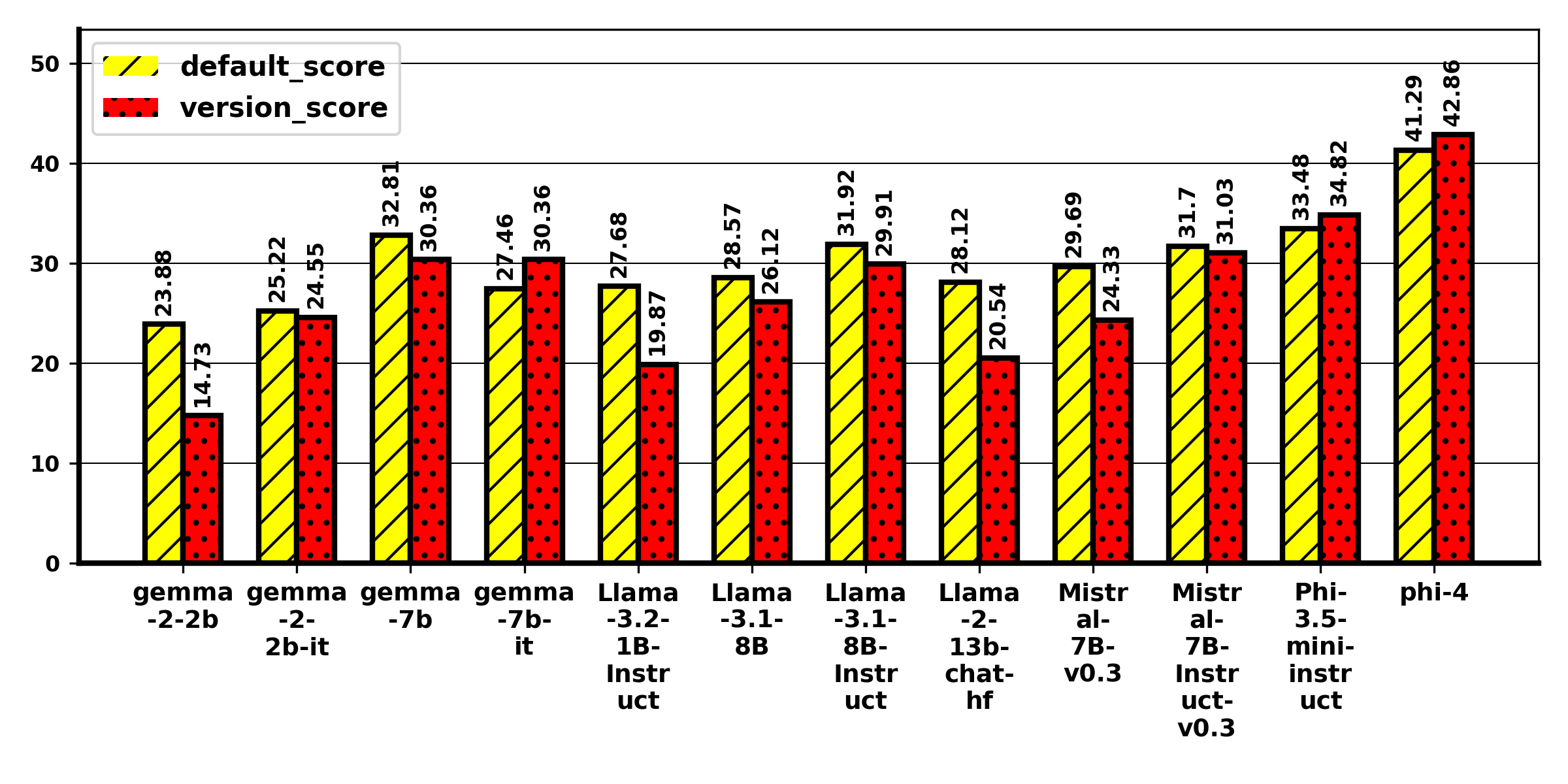}
    \caption{$\mathbf{v_{both}}$ results for GPQA benchmark across models.}
    \label{fig:results_1}
\end{figure}

\begin{figure}[h]
    \centering
    \includegraphics[width=\columnwidth]{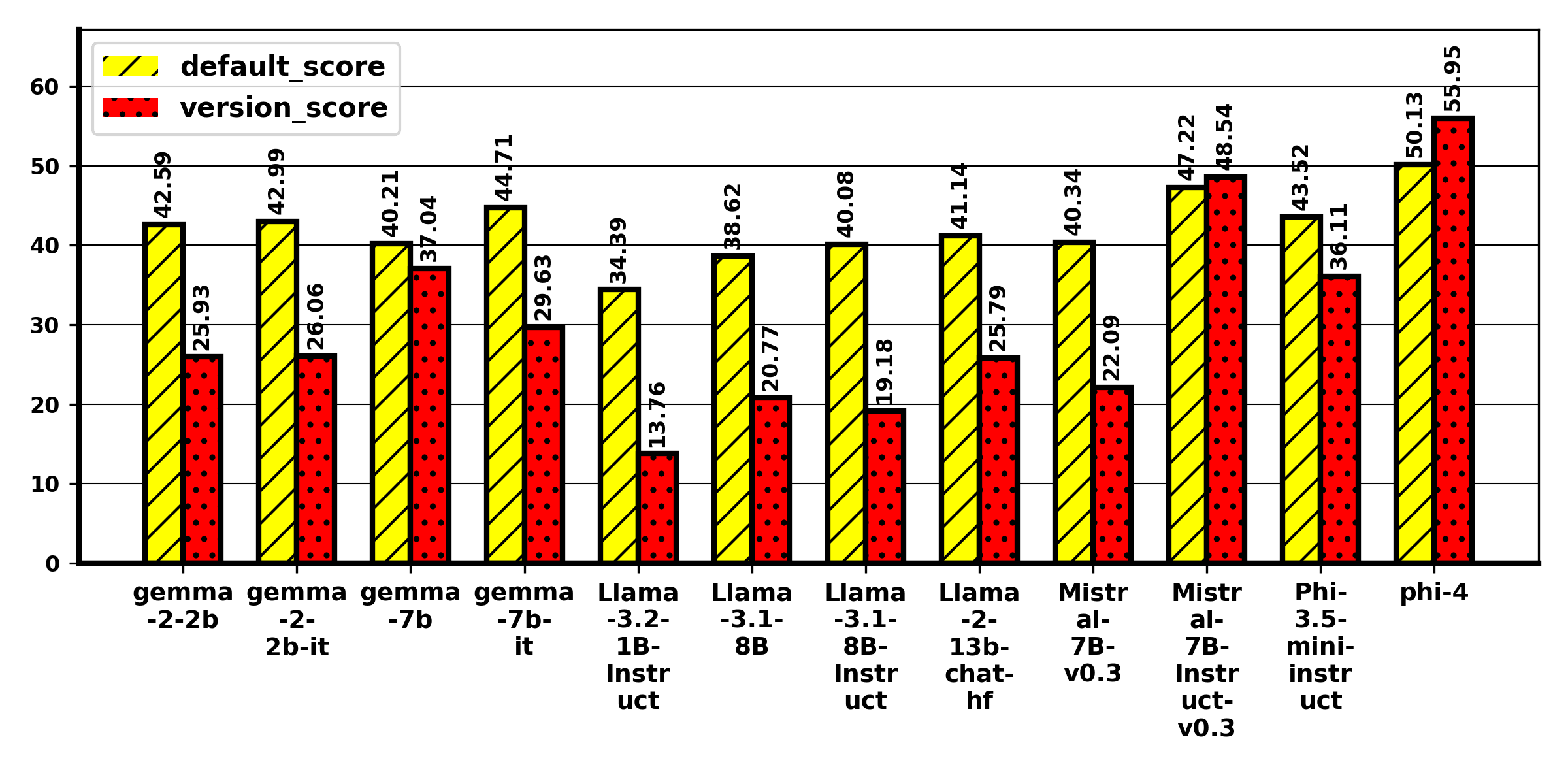}
    \caption{$\mathbf{v_{none}}$ results for MUSR benchmark across models.}
    \label{fig:results_1}
\end{figure}

\begin{figure}[h]
    \centering
    \includegraphics[width=\columnwidth]{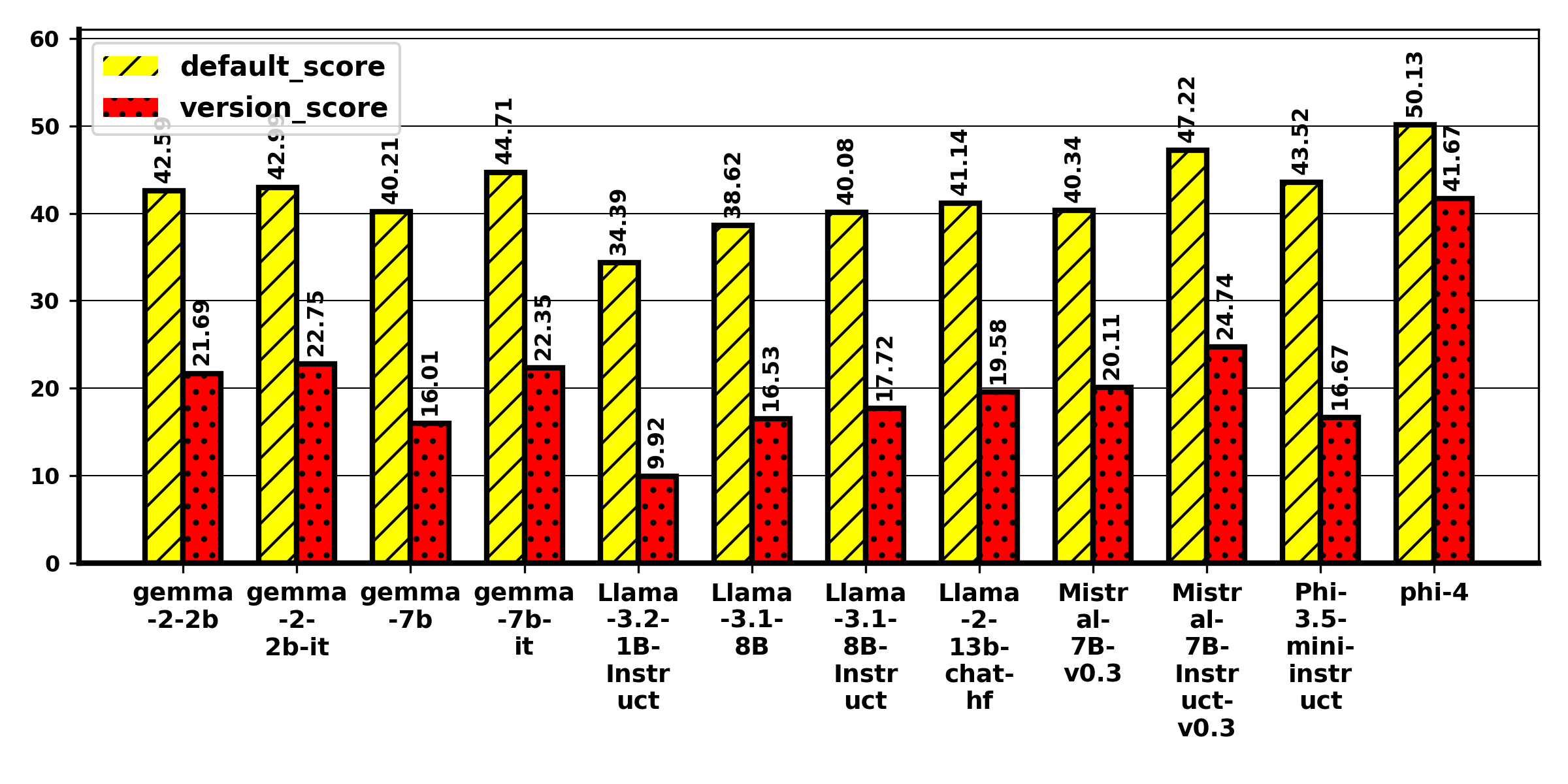}
    \caption{$\mathbf{v_{all}}$ results for MUSR benchmark across models.}
    \label{fig:results_1}
\end{figure}

\begin{figure}[h]
    \centering
    \includegraphics[width=\columnwidth]{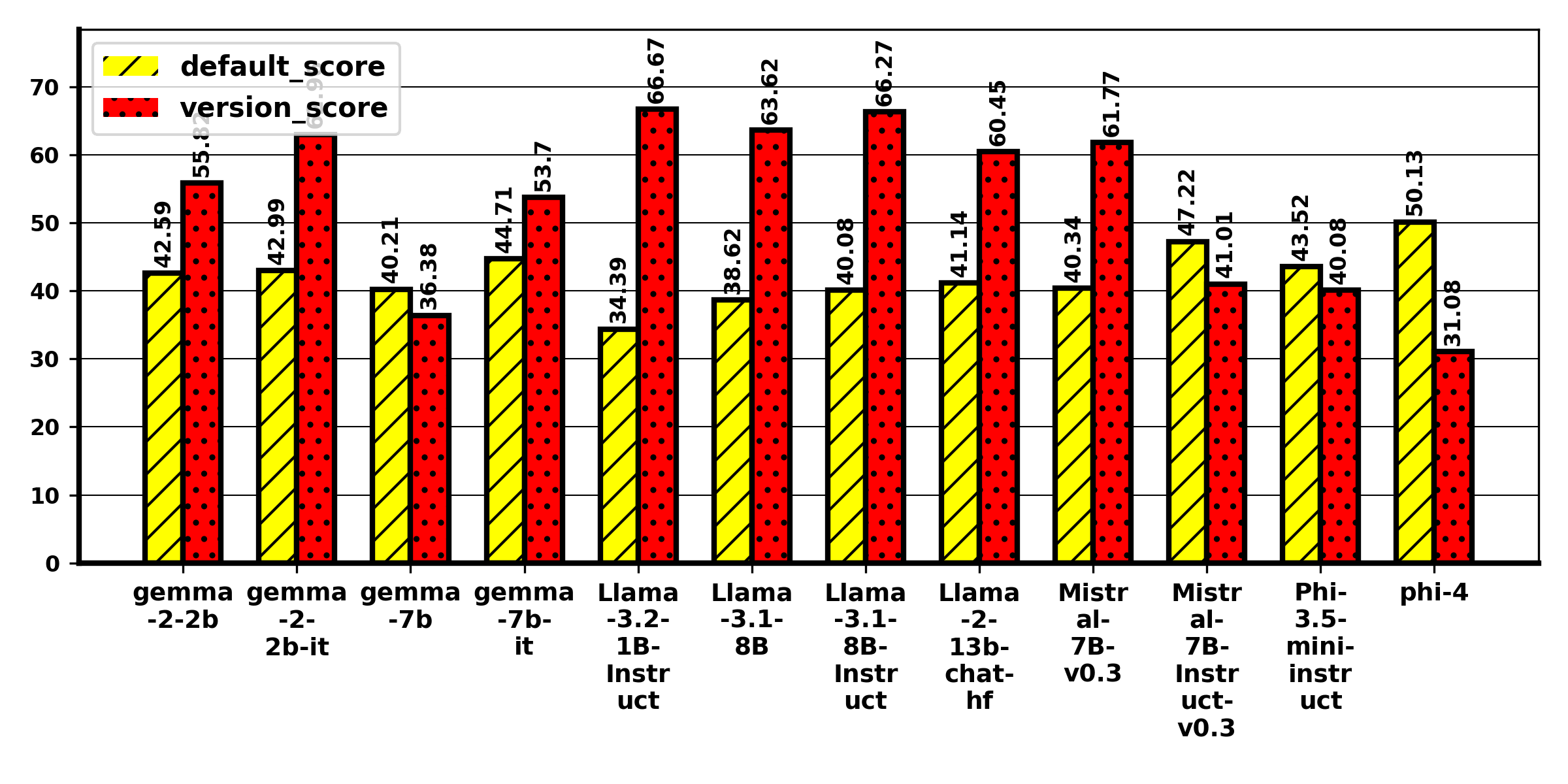}
    \caption{$\mathbf{v_{none-correct}}$ results for MUSR benchmark across models.}
    \label{fig:results_1}
\end{figure}

\begin{figure}[h]
    \centering
    \includegraphics[width=\columnwidth]{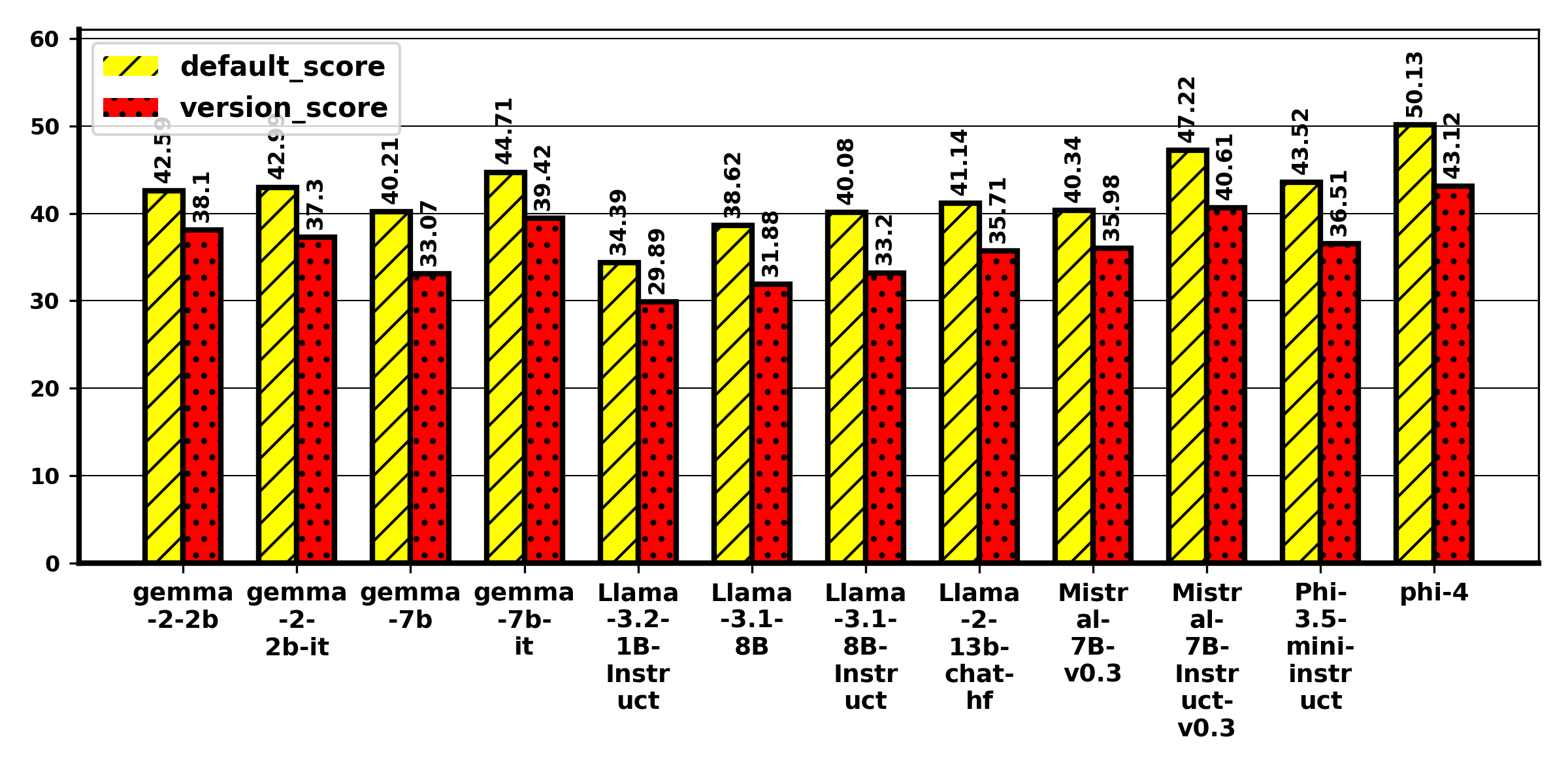}
    \caption{$\mathbf{v_{both}}$ results for MUSR benchmark across models.}
    \label{fig:results_1}
\end{figure}

\end{document}